%% file: main.tex
\documentclass[11pt]{article}

\usepackage[final]{acl}

\usepackage{times}
\usepackage{latexsym}

\usepackage[T1]{fontenc}

\usepackage[utf8]{inputenc}

\usepackage{microtype}

\usepackage{inconsolata}

\usepackage{graphicx}
\usepackage{tabularx}
\usepackage{makecell}
\usepackage{multirow}
\usepackage{booktabs}
\usepackage{algorithm}
\usepackage{caption}
\usepackage{algpseudocode}
\usepackage[table]{xcolor}
\usepackage{pifont}   
\definecolor{lightskyblue}{rgb}{0.53, 0.81, 0.98}
\newcommand{\cmark}{\textcolor{green!60!black}{\ding{51}}} 
\newcommand{\xmark}{\textcolor{red!70!black}{\ding{55}}}   

\usepackage{array}
\usepackage{amsmath}
\usepackage{amssymb}
\usepackage{kotex}
\usepackage{mathtools}
\usepackage{float}
\usepackage{pifont}

\title{LiteStage: Latency-aware Layer Skipping for Multi-stage Reasoning}

\author{Beomseok Kang, Jiwon Song, Jae-Joon Kim \\
  Seoul National University \\
  \texttt{\{beomseok,jiwon.song,kimjaejoon\}@snu.ac.kr}}

\begin{document}
\maketitle

\input{sec/abstract}
\input{sec/introduction}

\input{sec/related_works}
\input{sec/proposed_methods}

\input{sec/experimental_results}

\input{sec/conclusion}

\section*{Limitations}

LiteStage involves an offline profiling step to estimate accuracy-latency characteristics and select stage-wise layer budgets. While this profiling is performed once per model and setting and is amortized over repeated inference, it may introduce additional overhead compared to approaches that rely solely on fixed heuristics. In addition, generation early exit in our implementation is based on a simple confidence criterion, and more adaptive or task-specific exit strategies may further improve robustness. Our analysis also primarily focuses on short multi-stage reasoning, and the diagnostic results on deep reasoning suggest that the behavior under long-horizon generation can differ, indicating that extending LiteStage to such settings may require additional investigation. Moreover, our evaluation is concentrated on a three-stage multiple-choice QA pipeline over three benchmarks, and our transfer study covers only mild paraphrase-based distribution shifts. Evaluating LiteStage on other multi-stage workflows, stage formulations, prompt templates, and cross-dataset transfer remains future work. Finally, our robustness analysis varies evaluation seeds with a fixed fine-tuned multi-stage reasoning checkpoint; variance across training seeds is not measured.

\section*{Ethics Statement}
Although language models inherently present concerns regarding misuse, bias, and fairness, this work focuses solely on algorithmic and efficiency-oriented contributions. We do not foresee introducing any additional risks beyond those already associated with the base models.

We used LLMs in two limited ways. First, GPT-4.1-mini was used as a data transformation tool to generate paraphrased variants of the test questions for the distribution-shift evaluation (Appendix~\ref{sec:supp_dist_shift}); answer choices and labels were left unchanged, and the prompt is provided in full. Second, LLMs were used to polish the grammar and wording of text written by the authors. LLMs were not used in developing research ideas, designing methodologies, or analyzing results, and no scientific content or experimental descriptions were produced by LLMs.

\section*{Acknowledgements}

This work was supported in part by Institute of Information \& communications Technology Planning \& Evaluation (IITP) grant funded by the Korea government (MSIT) (No. RS-2026-25507427: Development of Efficient Architectures and Training Techniques for High-Performance Lightweight AI Models, No. RS-2025-02273157: Development of Low Power Training/Inference Technologies based on AI Semiconductors, No. RS-2026-25549926: AI Semiconductor Innovation Research Center), and by the Sejong Fellowship of National Research Foundation of Korea (NRF) funded by the Korea government (MSIT) (No. RS-2025-00561953: Efficient Deep Learning and Inference Algorithms for On-Device AI) (Corresponding Author: Jae-Joon Kim).

\clearpage
\bibliography{custom}

\clearpage
\input{sec/appendix}

\end{document}

%% file: sec/abstract.tex
\begin{abstract}
Multi-stage reasoning has emerged as an effective strategy for improving the reasoning capability of small language models by decomposing complex problems into sequential sub-stages. However, this comes at the cost of increased latency. We observe that existing adaptive acceleration techniques, such as layer skipping, struggle to balance efficiency and accuracy in this setting due to two key challenges: (1) stage-wise variation in skip sensitivity, and (2) extra generation of output tokens. To address these, we propose LiteStage, a latency-aware layer skipping framework for multi-stage reasoning. LiteStage combines a stage-wise offline search that allocates optimal layer budgets with an online confidence-based generation early exit to suppress redundant decoding. Experiments on three benchmarks, \textit{e.g.}, OBQA, CSQA, and StrategyQA, show LiteStage outperforms prior training-free layer skipping methods. Code: \textcolor[HTML]{3D7FD9}{\url{https://github.com/beomseokg/LiteStage}}.
\end{abstract}

%% file: sec/introduction.tex
\section{Introduction}

\begin{figure}[t]
\centering
\includegraphics[width=0.95\linewidth]{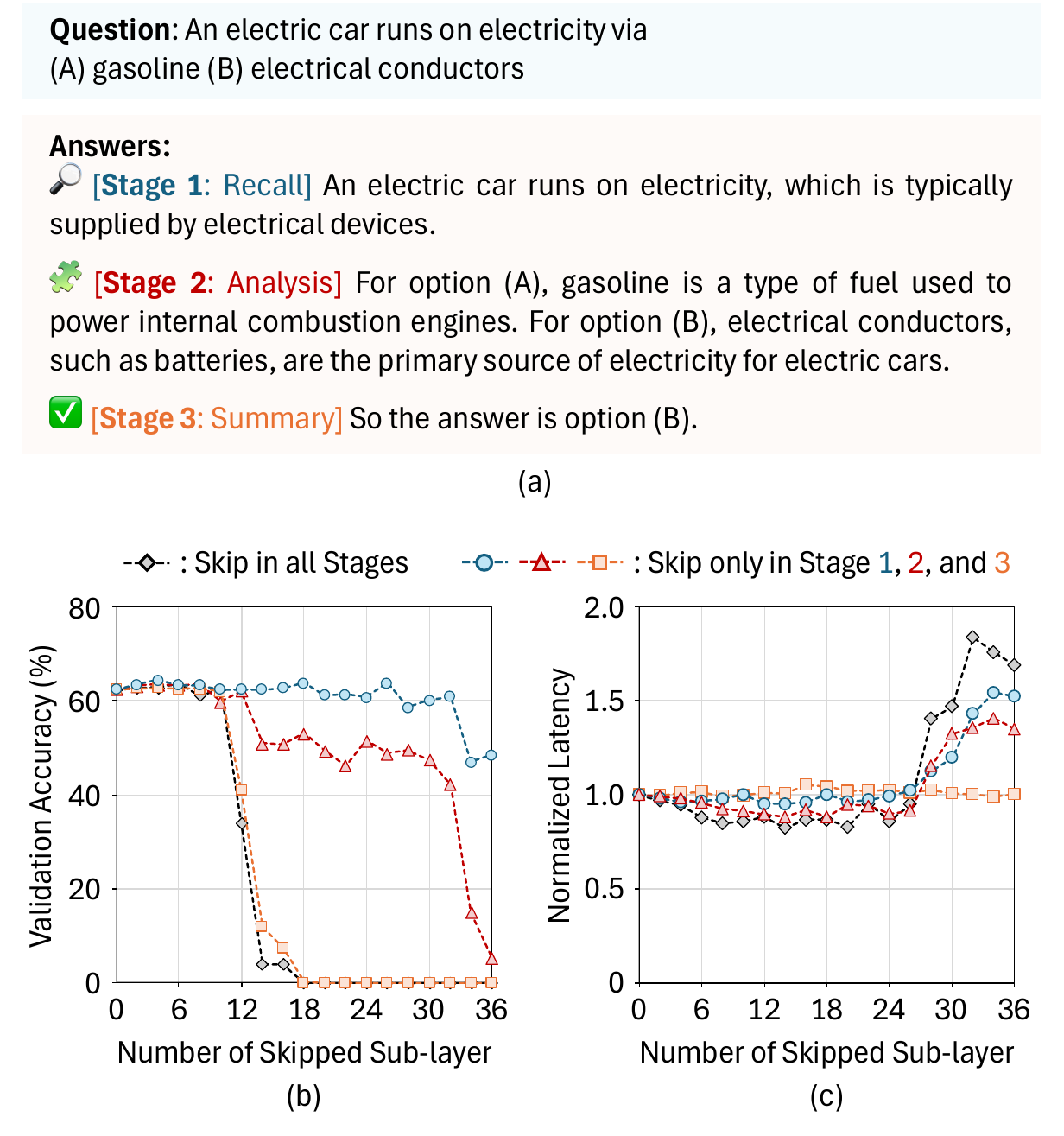}
\vspace{-2mm}
\caption{\textbf{Multi-stage Reasoning.} (a) An example of multi-stage reasoning introduced in TinyThinker~\cite{piao2024tinythinker}. The process comprises three stages: Stage~1~(\emph{Recall}) generates an initial solution idea, Stage~2~(\emph{Analysis}) evaluates candidate options through explicit reasoning, and Stage~3~(\emph{Summary}) produces the final conclusion. (b)-(c) Accuracy and latency profiles when layer skipping is applied to a \emph{single stage} while keeping the remaining stages at full depth.} 

\vspace{-4.5mm}
\label{figure:intro}
\end{figure}

Recent research on reasoning spans a broad spectrum, ranging from deep reasoning with long-horizon generation~\cite{hongru2025self,jin2024impact} to short multi-stage reasoning~\cite{wang2025stepwise, yang2025markov}. While the former has attracted significant attention, many practical question answering and decision-making tasks fall into the latter category~\cite{chen2024not}. In such settings, problems are decomposed into a few structured stages such as knowledge recall, local inference, and decision aggregation~\cite{piao2024tinythinker} (see Figure~\ref{figure:intro}(a)), allowing them to be solved without long, deeply nested derivations. 


\begin{table*}[t]
\centering
\renewcommand{\arraystretch}{1.0}
\setlength{\tabcolsep}{5pt}
\fontsize{9pt}{9pt}\selectfont

\caption {\textbf{Comparison of Key Features.} LiteStage is a training-free, stage-wise optimization strategy that adaptively allocates layer budgets across reasoning stages, unlike methods that either apply static layer skipping (training-free) or require additional training for dynamic skipping (\textit{e.g.}, routers).}

\vspace{-2mm}

\begin{tabular}{l|cccccc}

\toprule

\multirow{2}{*}{\textbf{Method / Contribution}} &
Training- & Latency- & Concise & Sub-layer & Adaptive & Importance \\
 & free & aware & Generation & Skipping & Skipping & Metric \\

\midrule

LayerSkip~\cite{elhoushi2024layerskip} & \xmark & \xmark & - & \xmark & \cmark & depth \\
MoD~\cite{raposo2024mixture} & \xmark & \xmark & - & \xmark & \cmark & router \\
SkipDecode~\cite{del2023skipdecode} & \cmark & \xmark & \xmark & \xmark & \cmark & heuristic \\
UnifiedSkip~\cite{liu2024accelerating} & \cmark & \xmark & \xmark & \xmark & \xmark & heuristic \\
AdaSkip~\cite{he2025adaskip} & \cmark & \xmark & \xmark & \cmark & \xmark & cosine \\
\rowcolor{lightskyblue!30} \textbf{LiteStage~(Ours)} & \cmark & \cmark & \cmark & \cmark & \cmark & cosine \\ %

\bottomrule

\end{tabular}
\label{table:related_works}
\vspace{-3mm}
\end{table*}


This form of \emph{short multi-stage reasoning} is particularly common in small language models, whose limited capacity often prevents reliable single-step deep reasoning~\cite{li2024teaching}. However, executing multiple reasoning stages incurs non-trivial latency~\cite{kim2024llm}, so inference can remain slow even for lightweight models. Applying existing acceleration techniques therefore appears natural, but multi-stage inference introduces unique challenges. Reasoning stages vary substantially in decoding length and token diversity, leading to non-uniform information density across stages~\cite{dai2024improve}. As a result, some stages tolerate aggressive acceleration, while others are highly sensitive, limiting achievable accuracy. Effective acceleration thus requires adapting efficiency to the distinct computational demands of each reasoning stage rather applying uniform compression.

Adaptive computation in LLMs has been widely explored through \textit{layer skipping} \cite{raposo2024mixture, men2024shortgpt, he2025adaskip}, which reduces computation by bypassing redundant layers. However, determining how many layers to skip at each reasoning stage remains challenging. As shown in Figure~\ref{figure:intro}(b), the degree of accuracy degradation differs across stages: Stage~1 is notably robust to layer skipping, whereas other stages are more sensitive. In contrast, latency benefits are most pronounced in Stage~2 (see Figure~\ref{figure:intro}(c)), due to its longer generation (\textbf{asymmetric trade-offs}). Moreover, approximate decoding often induces models to generate more tokens, which can result in even slower end-to-end inference than the full-layer, \textit{i.e.}, normalized latency > 1.0, despite reduced per-token latency (\textbf{extra generation}).

In this paper, we present \textbf{LiteStage}, a latency-aware layer skipping framework for multi-stage reasoning in small LLMs. LiteStage comprises two complementary components, an offline configuration and an online adjustment, that jointly address the key challenges. 
In the \textbf{offline phase}, LiteStage searches for \textit{layer budget}, \textit{i.e.}, the number of layers to skip, for each stage to minimize latency under an accuracy constraint. This search proceeds from the longest stage to the shortest, effectively accelerating slow stages while preventing accuracy collapse. In the \textbf{online phase}, LiteStage addresses the underexplored side effect of layer skipping, longer generation, by monitoring token confidence during decoding and \textit{terminating generation early} when confidence falls below a threshold, avoiding redundant generation. LiteStage's key features are summarized in Table~\ref{table:related_works}.

Our technical novelty lies in addressing the timely problem of accelerating reasoning in small LLMs. In multi-stage reasoning, we show stage-heterogeneous sensitivity that breaks the uniform-compression assumption of prior work. We also reveal an efficiency paradox where aggressive skipping increases decoding length, limiting end-to-end speedup. These observations motivate stage- and generation-aware acceleration for reasoning. LiteStage addresses these two phenomena with a co-design: stage-wise budget allocation resolves the stage heterogeneity, generation early exit resolves the efficiency paradox, and the two expand each other's feasible region: without early exit, aggressive skipping triggers extra generation that shrinks the viable budget search space, and without stage-wise allocation, early exit cannot exploit stage heterogeneity (Section~\ref{sec:early_exit}).

%% file: sec/related_works.tex
\section{Related Works}

\noindent\textbf{Multi-stage Generation.}
Recent works on reasoning tasks employ various forms of multi-stage inference. TinyThinker~\cite{piao2024tinythinker} introduces a deductive recall--analysis--summary reasoning cycle, showing progressive accuracy gains. DeAR~\cite{xue2024decompose} adopts a decomposition--analysis--rethinking process, refining intermediate answers across stages. CasCoD~\cite{dai2024improve} distills decomposed chain-of-thoughts to enhance small-model reasoning. Self-Discover~\cite{zhou2024self} dynamically organizes reasoning structures for each problem. LLaVA-CoT~\cite{xu2025llava} extends this idea to multi-modal settings. However, these works focus on reasoning quality rather than computational efficiency. 

\noindent\textbf{Layer Skip.} Layer-skipping methods fall into: (1) \textit{training-based} and (2) \textit{training-free} approaches. Early works like LayerSkip~\cite{elhoushi2024layerskip}, DeeBERT~\cite{xin2020deebert}, and EE-LLM~\cite{chen2023ee} perform early exit from intermediate layers \cite{fan2024not}. Recent router-based methods including Mixture-of-Depth~\cite{raposo2024mixture} skip layers dynamically but require additional training, while layer studies~\cite{luo2025diffskip, he2024router, luo2025adaptive, bae2025mixture} reduce cost by fine-tuning only routers. These training-based approaches are orthogonal to LiteStage: since our stage-wise budget allocation and generation early exit operate only on the decoding behavior of a given model, they can in principle be applied on top of LayerSkip-trained checkpoints or MoD routers, which we leave for future work.

Multi-stage reasoning models are often distilled from curated reasoning paths generated by larger models, which are not always available. Thus, \textit{training-free} acceleration is more practical in our setting. Representative methods include SkipDecode~\cite{del2023skipdecode}, which progressively skips more layers toward deeper decoding layers; Unified Skipping~\cite{liu2024accelerating}, which periodically skips layers (\textit{e.g.}, layers 3, 6, 9); and ShortGPT~\cite{men2024shortgpt}, which uses cosine similarity to estimate block importance. AdaSkip~\cite{he2025adaskip} further estimates importance at the sub-layer level, but these methods typically apply temporally uniform skipping policies, leading to suboptimal accuracy-efficiency trade-offs.

\noindent\textbf{Generation Early Exit.} 
Prior works mainly address redundant explanations in long reasoning traces. \citet{reasoningknow} trains probing heads to estimate confidence on intermediate answers, while training-free methods typically rely on heuristics. ES-CoT~\cite{escot} stops when answers repeat, and logit-based methods~\cite{deer,entropy} monitor confidence or entropy when \textit{</think>} token appears. These approaches focus on deep reasoning models and do not address extra generation caused by model compression.

%% file: sec/proposed_methods.tex
\section{Proposed Methods}
\label{sec:proposed_methods}

\noindent\textbf{Problem Statement.} Our primary objective is to search the stage-wise layer budget $\mathbb{L}$, \textit{i.e.}, a set of sub-layer indices to skip for each stage, that produces the minimal latency within a given accuracy threshold $\epsilon$. Formally, the objective is given as:
\begin{equation}
\arg\min_{\mathbb{L}}
\mathbb{E}_{d\sim\mathbb{D}}
\left[
\mathcal{T}(\mathcal{M}_{\mathbb{L}}(d))
\right]
\;\text{s.t.}\;
\mathcal{A}(\mathcal{M}_{\mathbb{L}})
\ge
\mathcal{A}(\mathcal{M})-\epsilon
\small
\label{equation:objective}
\end{equation}

\noindent where $\mathcal{T}$ and $\mathcal{A}$ denote the inference latency and accuracy of the model; $\mathcal{M}_{\mathbb{L}}$ and $\mathcal{M}$ represent models with layer skipping under the layer budget $\mathbb{L}$ and full layers, respectively; and $\mathbb{D}$ is a test dataset.



\subsection{Overview of LiteStage}

The details of each component in LiteStage are described in the following sections, and an overview is shown in Figure \ref{figure:overview}. The framework consists of two phases: (1) an \textit{offline configuration} that determines the optimal set of sub-layers to skip at each stage, and (2) an \textit{online adjustment} that compensates for inefficiencies during generation.

In the offline phase, we first estimate layer importance using cosine similarity to define priority of removable layers (\textbf{Step 1}). We then perform a greedy search that determines the \textit{layer budget} (\textit{i.e.}, the number of layers to skip) for each stage, from the longest reasoning stage to the shortest, to minimize latency within an accuracy threshold (\textbf{Step 2}). In the online phase, we address inefficiencies caused by extended generation. We observe that layer skipping often produces extra tokens with lower confidence, and thus combine it with confidence-based generation early exit (\textbf{Step 3}).



\subsection{Step 1: Estimate Layer Importance} 

We first need a layer skipping policy for selecting which layers to skip given a target skip count. We adopt cosine similarity as a proxy for layer importance \cite{men2024shortgpt, he2025adaskip}.

\begin{figure*}[t]
\centering
\includegraphics[width=0.95\linewidth]{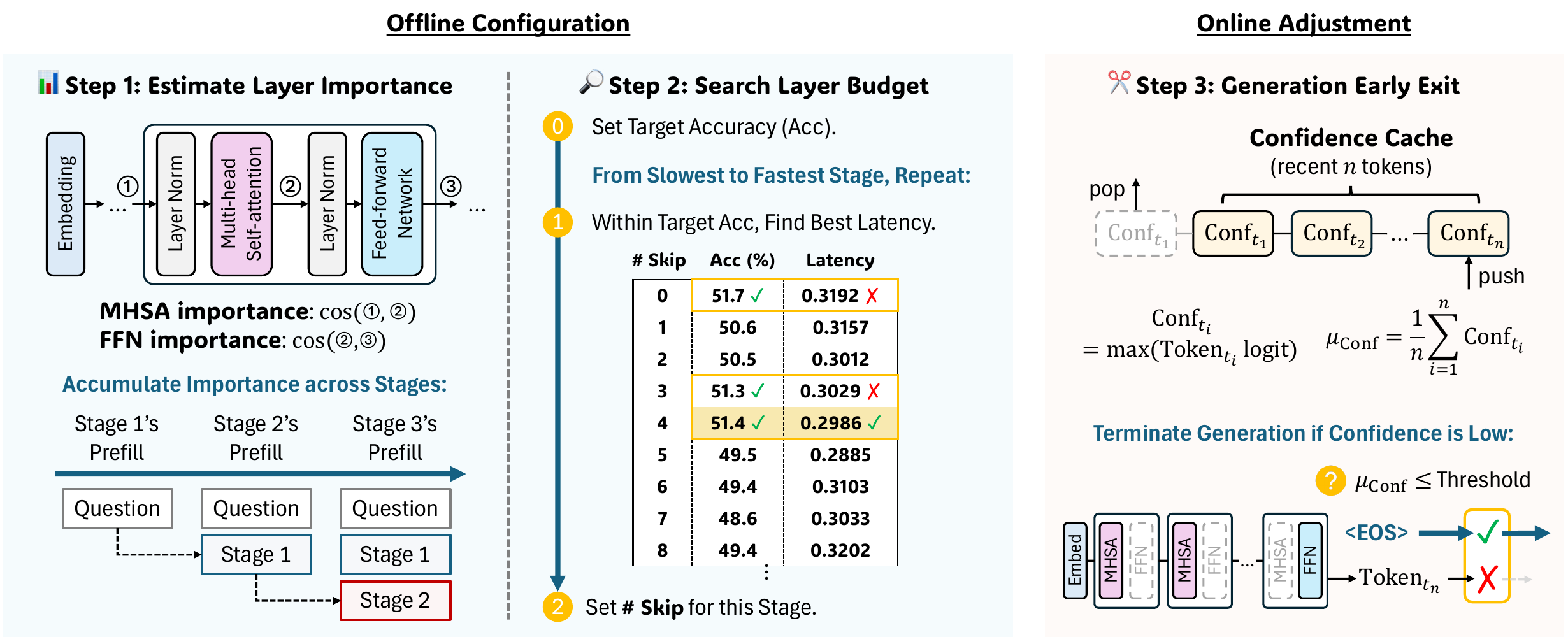}
\caption{\textbf{Overview of LiteStage.} The proposed method consists of an \textit{offline configuration} (Steps 1–2) and an \textit{online adjustment} (Step 3). In the offline phase, (1) layer importance is estimated at the sub-layer level (MHSA and FFN) and accumulated across stages, followed by (2) a search for the optimal layer budget that minimizes latency within a target accuracy. In the online phase, (3) \textit{generation early exit} dynamically terminates decoding when the average confidence of recent tokens falls below a threshold, preventing excessive generation length and ensuring consistent efficiency gains.}
\vspace{-2mm}
\label{figure:overview}
\end{figure*}

\noindent\textbf{Skipping at Sub-Layer Granularity.} We estimate layer importance at the sub-layer level, following the approach of AdaSkip \cite{he2025adaskip}. This finer granularity enables us to independently assess the influence of multi-head self-attention (MHSA) and feed-forward network (FFN) sub-layers. Our contribution lies \textit{not} in designing a new importance proxy; we adopt the existing one to rank sub-layers.

Specifically, the importance of each sub-layer is estimated as follows:
\vspace{-2mm}
{\small
\begin{align}
    I_{\text{MHSA}}^{(j)} &= 1 - \frac{1}{N} \sum_{n=0}^{N-1} \cos(\text{MHSA}^{(j)}(x)+x, x) \label{equation:att_importance} \\
    I_{\text{FFN}}^{(j)} &= 1 - \frac{1}{N}\sum_{n=0}^{N-1}\cos(\text{FFN}^{(j)}(x)+x, x) \label{equation:mlp_importance}
\end{align}}

\noindent where $I_\text{MHSA}^{(j)}$ and $I_\text{FFN}^{(j)}$ denote the importance of the $j$-th MHSA and FFN layers, respectively. We compute cosine similarity between input and output of each sub-layer, as described in Figure \ref{figure:overview} (Step 1), during prefilling and average over $N$ validation samples. Equations (\ref{equation:att_importance}) and (\ref{equation:mlp_importance}) are also averaged across stages, though omitted here for clarity. Higher similarity indicates that input and output representations are more redundant, \textit{i.e.}, less important sub-layer. We empirically validate this proxy and its validation-to-test transferability in Appendix~\ref{sec:supp_importance}. 

This estimation is first performed at Stage~1 using the corresponding prompt, and is subsequently accumulated in Stages~2 and~3 by computing the cosine similarities with their respective prompts. 
This process is conducted offline and only once per dataset. We report $I_\text{MHSA}$ and $I_\text{FFN}$ on evaluation datasets in Figure~\ref{figure:supp_importance} (Appendix), where we observe little variation across datasets.

\subsection{Step 2: Search Layer Budget} The first core component of LiteStage is the search for the optimal stage-wise layer budget. To optimize latency under a fixed accuracy constraint, we first construct an accuracy-latency profile by varying the layer budget in the \textit{longest} reasoning stage using validation data, while keeping all other stages full-layer. This prioritizes accelerating the longest stage, which most effectively reduces the overall end-to-end latency. We then progressively explore layer skipping in the remaining stages. Please refer to Appendix~\ref{sec:supp_dist_shift} for a discussion on the robustness of our validation-based search method under test sets with mild distribution shifts.

\noindent\textbf{Skipping from the Longest Stage.} Figure \ref{figure:overview} (Step 2) illustrates this process using the actual accuracy-latency profile of Stage 2, which is the longest stage, on CSQA with a TinyLlama 1.1B model (see Table~\ref{table:supp_data_statistics} (Appendix) for data statistics). Let us consider an accuracy threshold ($\epsilon$) of 1.0\% as an example. Given that the baseline accuracy without layer skipping ("\# Skip": 0) is 51.7\%, the target accuracy becomes 50.7\% (\textit{i.e.}, 51.7\%-1.0\%). The layer budgets that satisfy this target accuracy are highlighted with orange table borders. Among these, skipping four layers ("\# Skip": 4) yields the lowest latency of 0.2986, representing the optimal layer budget. This completes a single greedy search iteration. For the next longest stages, we repeat this profiling process in the same manner, but with the previously optimized stages already under their selected layer budgets (\textit{e.g.}, applying "\# Skip": 4 for Stage 2). We maintain the same target accuracy of 50.7\%, allowing most of the accuracy degradation to occur in the first search, which is desirable since the longest stage will be effectively accelerated. This example is for illustration only and does not imply that $\epsilon$ is fixed to 1\%; in practice, we use multiple $\epsilon$ values.

\noindent\textbf{Why Accuracy-Latency Profiling Matters.} Despite its simplicity, search-based layer budget allocation provides several key advantages, that are not captured when relying solely on cosine similarity.

(1) \textit{avoiding sub-optimal latency}: we often observe that accuracy and latency do not change monotonically with the number of skipped layers. As illustrated in Figure~\ref{figure:overview}, beyond the "\# Skip" of 5, the latency gain diminishes, and further skipping can even increase end-to-end inference time. This counterintuitive behavior arises because, although per-token latency is reduced, aggressive approximation often increases the total number of generated tokens, resulting in higher overall latency. Since our search procedure jointly optimizes accuracy and latency, such configurations are naturally pruned.

(2) \textit{interaction between stages}: the profile ensures that interactions between stages are explicitly considered during the budget search. 
We perform stage-wise budget allocation sequentially from the longest stage (Stage 2) to the shortest (Stage 3). When searching for Stage 2’s layer budget, the other stages use full layers. Once Stage 2’s budget is fixed, Stage 1’s budget is searched while Stage 2 operates under skipping. Finally, Stage 3’s budget is determined while both Stage 1 and Stage 2 use their selected skipping policies. This is different from an “independent search”, where Stage $n$’s budget would be determined while all other stages remain full-layer.

(3) \textit{identifying most sensitive stage}: as shown in Figure \ref{figure:intro}(b), uniformly skipping layers across all stages yields an accuracy profile nearly identical to skipping only in Stage~3. This indicates that the final accuracy is bottlenecked by the most sensitive stage. The profile explicitly reveals such sensitivity, enabling non-uniform allocation that protects critical stages (\textit{e.g.}, Stage~3), allowing more aggressive skipping in robust stages (\textit{e.g.}, Stages~1 and~2).

\begin{table}[t]
\centering
\renewcommand{\arraystretch}{0.9}
\setlength{\tabcolsep}{4pt}
\fontsize{8pt}{8pt}\selectfont

\caption{\textbf{Offline Search Cost.} Average evaluation latency (minutes) per skipping configuration. The total search time scales with the number of layers (\textit{e.g.}, 22 for TinyLlama-1.1B) and reasoning stages (\textit{e.g.}, 3 in our setup). Results are measured on a single A6000 GPU.}

\vspace{-2mm}

\begin{tabular}{l|ccc}

\toprule
\textbf{Model~\textbackslash~Benchmark} & \textbf{OBQA} & \textbf{CSQA} & \textbf{StrategyQA} \\
\midrule
TinyLlama-1.1B & 2.51 & 5.57 & 0.78 \\
Qwen2.5-0.5B & 1.89 & 4.23 & 0.68 \\
\bottomrule

\end{tabular}

\vspace{-3mm}
\label{table:search_cost}
\end{table}

\noindent\textbf{Offline Search Cost.} 
Table~\ref{table:search_cost} reports the offline search overhead. Finding the optimal setting requires exploring combinations of skipped layers and reasoning stages, but this one-time search is performed on a single GPU and typically finishes within a few hours, \textit{e.g.}, 2.8/6.1/0.9 hours on OBQA/CSQA/StrategyQA for TinyLlama-1.1B, and 1.5/3.4/0.5 hours for Qwen2.5-0.5B. The selected configuration is then reused for all inference, incurring no additional cost.

\subsection{Step 3: Generation Early Exit}\label{sec:early_exit} 
While Step 2 determines a balanced layer budget, it leaves long outputs unchanged. To further improve speedup, we reduce redundant output tokens to recover the latency gains from layer skipping. Our hypothesis is that tokens induced by skipping contribute little to the final reasoning outcome.

\begin{figure}[t]
\centering
\includegraphics[width=\linewidth]{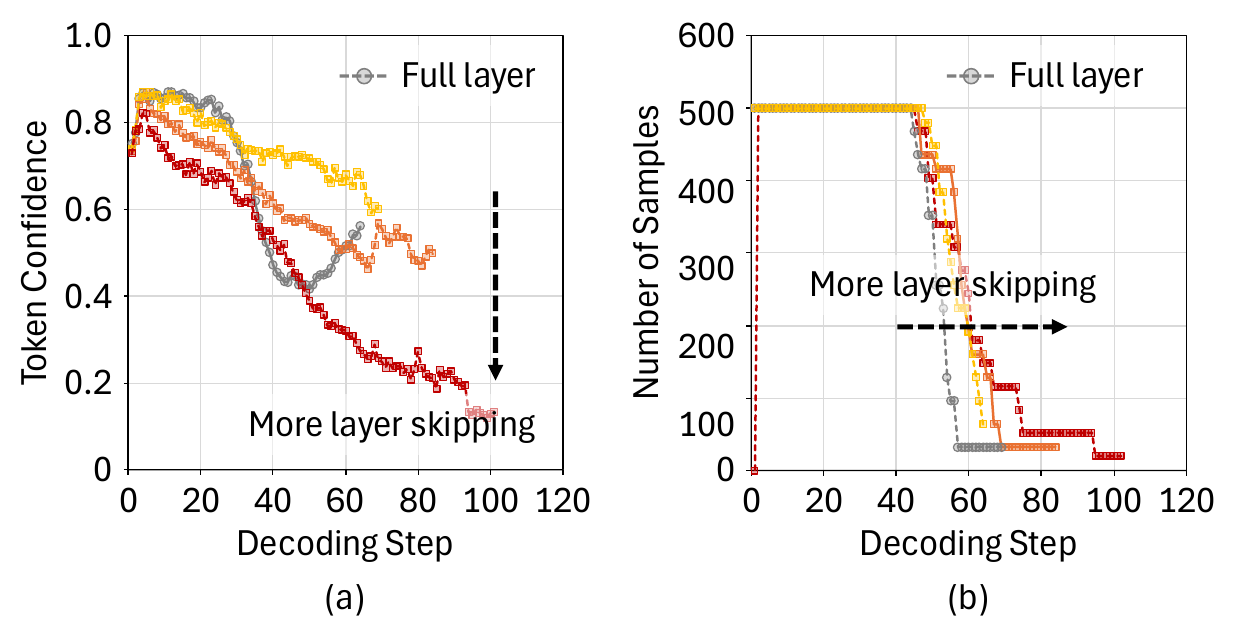}
\vspace{-7mm}
\caption{\textbf{Confidence of Output Tokens.} (a) Average token-level confidence across decoding steps under different layer-skipping levels. 
(b) Number of remaining samples during decoding (not yet exited), where more aggressive layer skipping leads to earlier exits due to confidence decay. Results are shown for skipping of 10 (yellow), 15 (orange), and 20 (red) sub-layers in Stage~1 on the OBQA validation set using TinyLlama-1.1B.}

\vspace{-3mm}
\label{figure:confidence}
\end{figure}

\noindent\textbf{Extra Tokens may not be Useful.} Figure \ref{figure:confidence} shows how token confidence evolves during decoding under four skip configurations, where confidence is defined as the maximum logit of each generated token. Confidence trajectories differ with layer skipping: skipped models show steadily decreasing confidence, whereas the full-layer model partially recovers in later steps. However, a common pattern remains that high confidence (${>}$0.5) mostly occurs in early decoding steps. This becomes clearer as more layers are skipped; for example, the red curve (20-layer skip) consistently exhibits lower confidence and longer generation than the yellow curve (10-layer skip). Based on this observation, we apply confidence-based early exit in generation, assuming that terminating these low-confidence tokens has little impact on accuracy.


\begin{algorithm}[t]
\caption{LiteStage: Stage-wise Layer Skipping with Early Exit}
\label{alg:litestage}
\begin{algorithmic}[1]

\Statex \textbf{Input:} Multi-stage pipeline $\{S_1,S_2,S_3\}$, model $M$, validation set $D_{\text{val}}$, accuracy threshold
$\varepsilon$, exit params $(n,\tau)$
\Statex \textbf{Output:} Skip configuration $\{L_1,L_2,L_3\}$

\Statex \textbf{Step 1: Layer-importance estimation}
\For{each stage $S_t$}
    \State Estimate sub-layer redundancy on $D_{\text{val}}$
    \State Rank sub-layers by importance
\EndFor

\Statex \textbf{Step 2: Stage-wise budget search}
\State Measure full-depth accuracy $A_0$, set $A_{\text{target}}=A_0-\varepsilon$
\For{stages in longest $\rightarrow$ shortest order}
    \For{candidate skip $k$}
        \State Apply skip to current stage only
        \State Evaluate accuracy $A(k)$ and latency $T(k)$ on $D_{\text{val}}$
    \EndFor
    \State Select fastest $k$ s.t. $A(k)\ge A_{\text{target}}$
\EndFor

\Statex \textbf{Step 3: Online confidence-based early exit}
\While{decoding}
    \If{moving-average confidence $<\tau$ for $n$ tokens}
        \State stop generation
    \EndIf
\EndWhile

\State \Return $\{L_1,L_2,L_3\}$

\end{algorithmic}
\end{algorithm}

\noindent\textbf{Confidence-based Termination.} Our approach is straightforward: as shown in Figure \ref{figure:overview} (Step 3), if the confidence of an output token falls below a threshold, we replace it with an end-of-sequence (EOS) token to stop generation. Because confidence can fluctuate across decoding steps, relying on a single token may cause premature termination. To stabilize the decision, we maintain a \textit{confidence cache} storing the most recent $n$ confidence values, and from the $n$-th step, compute the mean confidence $\mu_{\text{Conf}}$ to compare with a threshold. We set $n{=}5$ and the confidence threshold $\tau$ to 0.5 based on preliminary experiments; a sweep over $n$ and $\tau$ on OBQA (Appendix~\ref{app:exit_sweep}) shows that this default lies on a wide accuracy plateau for both TinyLlama-1.1B and Qwen2.5-0.5B.

We also incorporate generation early exit into the search process (Step~2) so that the accuracy-latency profile reflects real evaluation conditions. Without generation early exit, the effective search space can shrink when generation length becomes highly sensitive to skipping, causing latency gains to vanish beyond few skip levels. Generation early exit mitigates this, expanding the feasible search space and allowing more valid layer-skipping configurations.

We provide an algorithm-style description of LiteStage in Algorithm~\ref{alg:litestage} to improve clarity and facilitate understanding of the overall procedure.

%% file: sec/experimental_results.tex
\section{Experimental Results}
\label{sec:experimental_results}


\noindent\textbf{Dataset and Implementation.} 
We adopt the three-stage reasoning flow from TinyThinker~\cite{piao2024tinythinker} on three question-answering benchmarks: OpenBookQA (OBQA)~\cite{OpenBookQA2018}, CommonSenseQA (CSQA)~\cite{talmor2018commonsenseqa}, and StrategyQA~\cite{geva2021strategyqa}. Models are trained on TinyThinker’s reasoning paths before applying layer skipping. During validation and testing, reasoning paths are not accessible. 
Our method is primarily evaluated using TinyLlama-1.1B-Chat-v1.0~\cite{zhang2024tinyllama} and Qwen2.5-0.5B~\cite{qwen2025qwen25technicalreport}. Our training and evaluation procedures follow TinyThinker~\cite{piao2024tinythinker}, with adjusted hyperparameters. More details are available in Appendix~\ref{sec:supp_training_setups}.

\noindent\textbf{Baselines.} We consider recent training-free layer-skipping methods, SkipDecode \cite{del2023skipdecode}, UnifiedSkip \cite{liu2024accelerating}, and AdaSkip \cite{he2025adaskip}, as our baselines. 
Our main baseline is AdaSkip, as our layer-importance estimation also adopts their sub-layer-wise cosine similarity. We apply layer skipping only during decoding stages, and our implementation of the baselines also follows the AdaSkip's code\footnote{https://github.com/ASISys/AdaSkip}. We additionally compare with ShortGPT~\cite{men2024shortgpt} in Appendix~\ref{app:shortgpt}.

\begin{figure}[t]
\centering
\includegraphics[width=\linewidth]{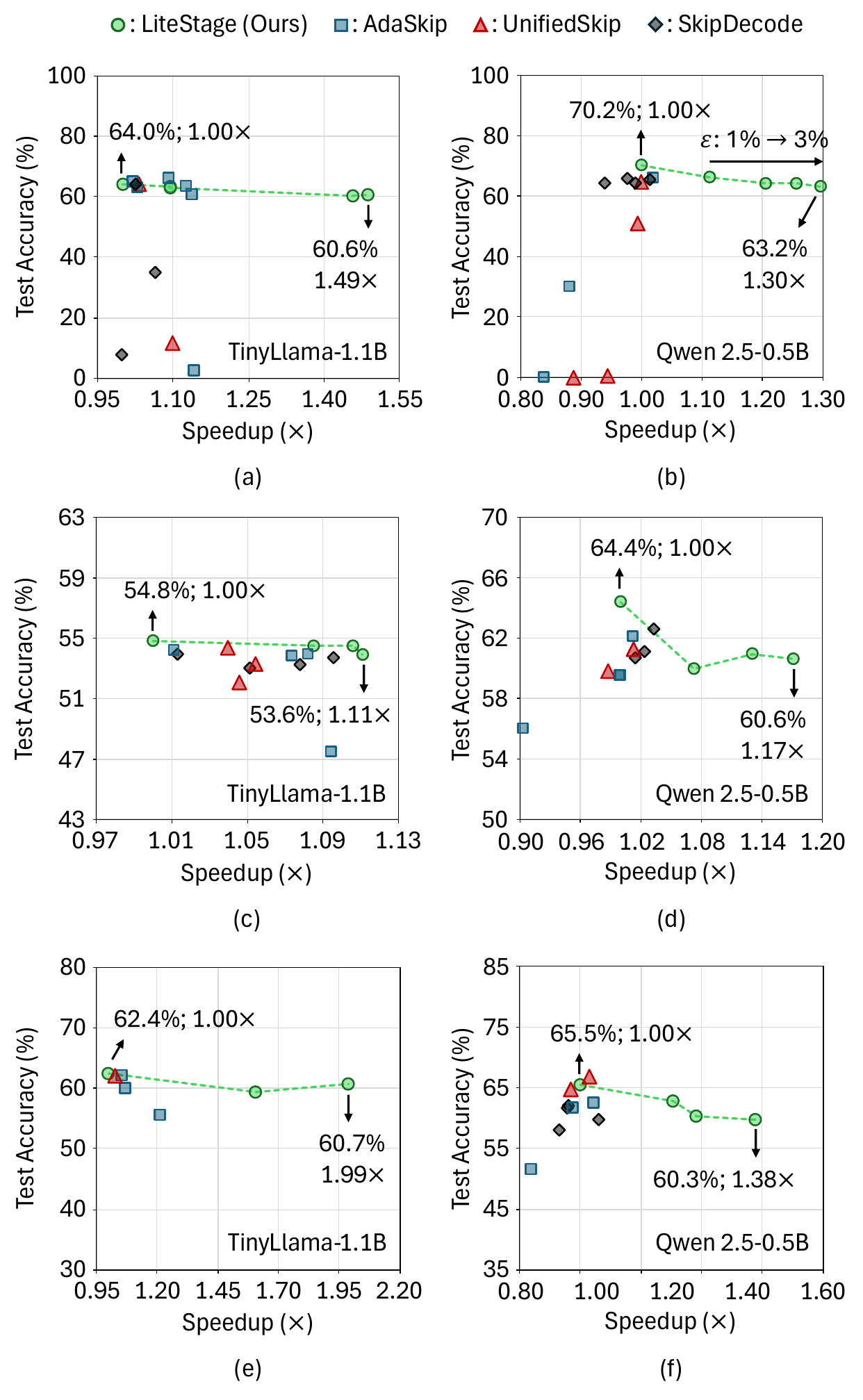}
\vspace{-7mm}
\caption{\textbf{Comparison with Baselines.} Accuracy–speedup trade-offs on the OBQA (row-1), CSQA (row-2), and StrategyQA (row-3) datasets, respectively. Results are experimented using TinyLlama-1.1B (col-1) and Qwen2.5-0.5B (col-2) models. The performance of the full-layer model is marked in the upper-left region of each plot (\textit{e.g.}, 64.0\% and 70.2\% for OBQA), and the speedup is normalized by the full-layer latency. Mean and standard deviation over three evaluation seeds at representative operating points are reported in Appendix~\ref{app:multiseed}.}

\vspace{-2mm}
\label{figure:pareto}
\end{figure}

\begin{table}[t]
\centering
\renewcommand{\arraystretch}{0.7}
\setlength{\tabcolsep}{4pt}
\fontsize{8pt}{8pt}\selectfont

\caption{\textbf{Non-uniform Layer Budget.} The number of skipped layers allocated to each stage by LiteStage across the three benchmarks. Lower rows represent more aggressive layer skipping with lower latency. Each row corresponds to a data point in Figure~\ref{figure:pareto}.}

\vspace{-2mm}

\begin{tabular}{l|ccc|ccc}

\toprule
& \multicolumn{3}{c|}{\textbf{TinyLlama-1.1B}} & \multicolumn{3}{c}{\textbf{Qwen2.5-0.5B}} \\
\midrule
& Stage 1 & Stage 2 & Stage 3 & Stage 1 & Stage 2 & Stage 3 \\
\midrule

\multirow{4}{*}{\rotatebox[origin=c]{90}
{OBQA}} & 5 & 3 & 4 & 6 & 0 & 1 \\
& 2 & 4 & 5 & 15 & 0 & 1 \\
& 21 & 4 & 4 & 15 & 1 & 1 \\
& 19 & 6 & 4 & 11 & 2 & 1 \\
\midrule


\multirow{3}{*}{\rotatebox[origin=c]{90}
{CSQA}} & 0 & 3 & 0 & 0 & 2 & 0 \\
& 1 & 3 & 5 & 1 & 2 & 1 \\
& 1 & 3 & 4 & 9 & 1 & 3 \\
\midrule


\multirow{3}{*}{\rotatebox[origin=c]{90}
{StrQA}} & 12 & 10 & 0 & 14 & 0 & 1 \\
& 0 & 15 & 0 & 3 & 5 & 0 \\
& - & - & - & 3 & 6 & 1 \\
\bottomrule

\end{tabular}
\label{table:layer_budget}
\vspace{-1mm}
\end{table}

\begin{figure}[t]
\centering
\includegraphics[width=\linewidth]{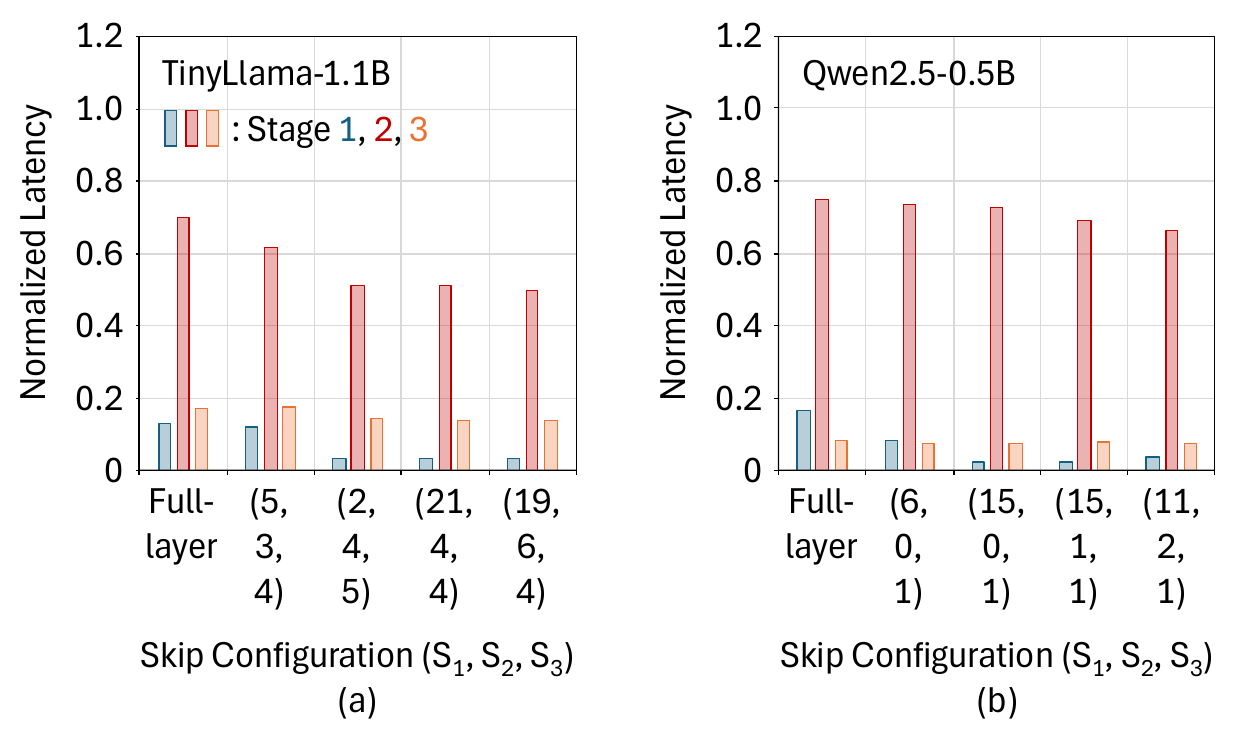}
\vspace{-7.5mm}
\caption{\textbf{Stage-wise Latency.} (a)-(b) show the normalized latency of each stage in TinyLlama-1.1B and Qwen2.5-0.5B, respectively, on OBQA. In the full-layer configuration, the sum of stage-wise latencies is normalized to 1.0. $S_1$, $S_2$, $S_3$ on the x-axis denotes the number of skipped layers at Stages 1, 2, and 3.}
\vspace{-2.5mm}
\label{figure:stage_latency}
\end{figure}

\subsection{Comparison with Baselines}
\label{sec:comparison}

Figure \ref{figure:pareto} compares LiteStage with three baselines on OBQA, CSQA, and StrategyQA. Notably, LiteStage preserves reasoning quality even at high acceleration ratios, where prior methods often degrade significantly. Baselines are evaluated by progressively increasing the number of skipped layers until speedup saturates. LiteStage consistently outperforms them, extending the achievable performance boundary. For example, in OBQA (Figure~\ref{figure:pareto}(a)), the strongest baseline AdaSkip matches our accuracy up to 1.10$\times$ speedup, but quickly collapses afterward, dropping to nearly 0\% accuracy, while LiteStage remains stable, achieving 1.49$\times$ speedup with 60.6\% accuracy. The achievable speedup depends on the user-specified $\epsilon$, which controls the accuracy–efficiency trade-off. In practice, we tune $\epsilon$ on a validation set (\textit{e.g.}, 1\%--3\% for Qwen2.5-0.5B), and select the configuration that best matches the accuracy-latency balance.

We highlight two key observations from these results: (1) how LiteStage mitigates severe accuracy degradation ($0\% {\rightarrow} 60\%$) via \textit{non-uniform layer budget}, and (2) extends the latency limit ($1.10\times{\rightarrow}1.49\times$) through \textit{generation early exit}.

\subsection{Benefits of Non-uniform Layer Budget} 
\label{sec:benefits1}

\noindent\textbf{Protection from Accuracy Collapse.} Table~\ref{table:layer_budget} presents layer budgets assigned by LiteStage for each stage. LiteStage consistently skips no more than five (TinyLlama) and three (Qwen) layers in Stage~3, reflecting its ability to limit compression in sensitive stages while accelerating more robust ones. This is because Stage~3 is notably more sensitive to skipping and often determines the overall accuracy (see Figures~\ref{figure:supp_ablation_tinyllama_obqa}--\ref{figure:supp_ablation_qwen_strategyqa}, Appendix). By allocating fewer skipped layers to such stages, LiteStage prevents accuracy collapse, demonstrating the effectiveness of stage-wise non-uniform skipping.


\noindent\textbf{Layer Budget Distribution.} Beyond protecting accuracy-sensitive stages, LiteStage adapts the layer budget across datasets. As shown in Figures~\ref{figure:intro}(b)-(c), Stage~1 is relatively robust to skipping, whereas Stage~2 dominates latency. Accordingly, when accuracy must be preserved, LiteStage allocates more skipping to Stage~1, as Stage~2 has little room for further skipping under the accuracy constraint. For example, on OBQA (Table~\ref{table:layer_budget}), most variation appears in the Stage~1 budget, increasing from $5{\rightarrow}21$ (TinyLlama) and $6{\rightarrow}15$ (Qwen).

As further speedup is required, LiteStage shifts budget to Stage~2 while reducing Stage~1. In the most aggressive OBQA setting (row 3 in Table~\ref{table:layer_budget}), budgets change from $21,4 {\rightarrow} 19,6$ (TinyLlama) and $15,1 {\rightarrow} 11,2$ (Qwen) for Stages~1 and~2. Thus, LiteStage first reduces Stage~1 latency and then Stage~2 when larger reductions are needed (Figure~\ref{figure:stage_latency}). A similar trend appears on StrategyQA, while on CSQA Stage~2 is more sensitive, so larger speedups come from further accelerating Stage~1.

\begin{figure}[t]
\centering
\includegraphics[width=\linewidth]{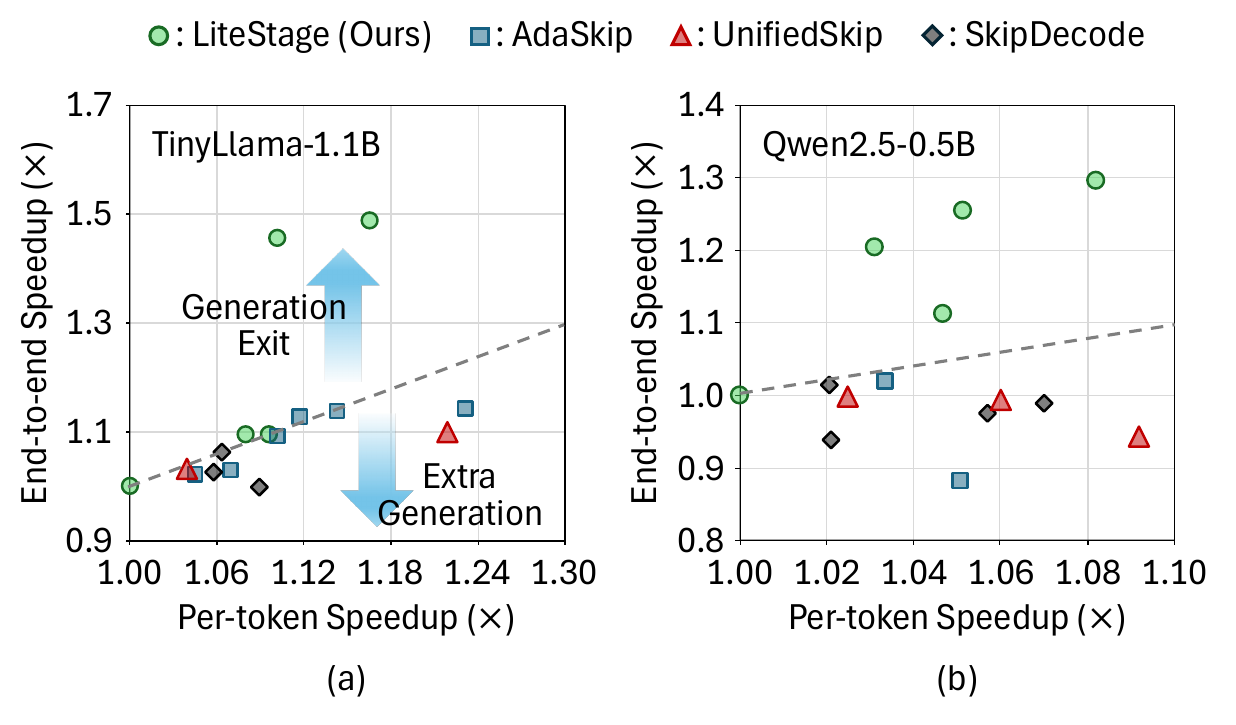}
\vspace{-7.5mm}
\caption{\textbf{Per-token and End-to-End Speedup.} (a)-(b) show the per-token and end-to-end speedup of TinyLlama-1.1B and Qwen2.5-0.5B, respectively, on the OBQA dataset. The full-layer baseline is normalized to a per-token and end-to-end speedup of 1.0. The gray dotted line denotes the $y{=}x$ reference.}
\vspace{-2.5mm}
\label{figure:per_token}
\end{figure}

\subsection{Benefits of Generation Early Exit}
\label{sec:benefits2}

\noindent\textbf{Per-token vs End-to-End Speedup.} Extended generation hinders end-to-end speedup, even though layer skipping reduces per-token latency. Figure~\ref{figure:per_token} compares per-token and end-to-end speedup, where the ideal case follows the dotted line ($y{=}x$), \textit{i.e.}, per-token speedup translates to end-to-end speedup. For the baselines, per-token speedup keeps increasing, but end-to-end speedup saturates or even drops, especially under aggressive skipping. In contrast, with generation early exit, LiteStage improves end-to-end speedup, often exceeding per-token speedup and fully benefiting from layer skipping.

\begin{figure}[t]
\centering
\includegraphics[width=\linewidth]{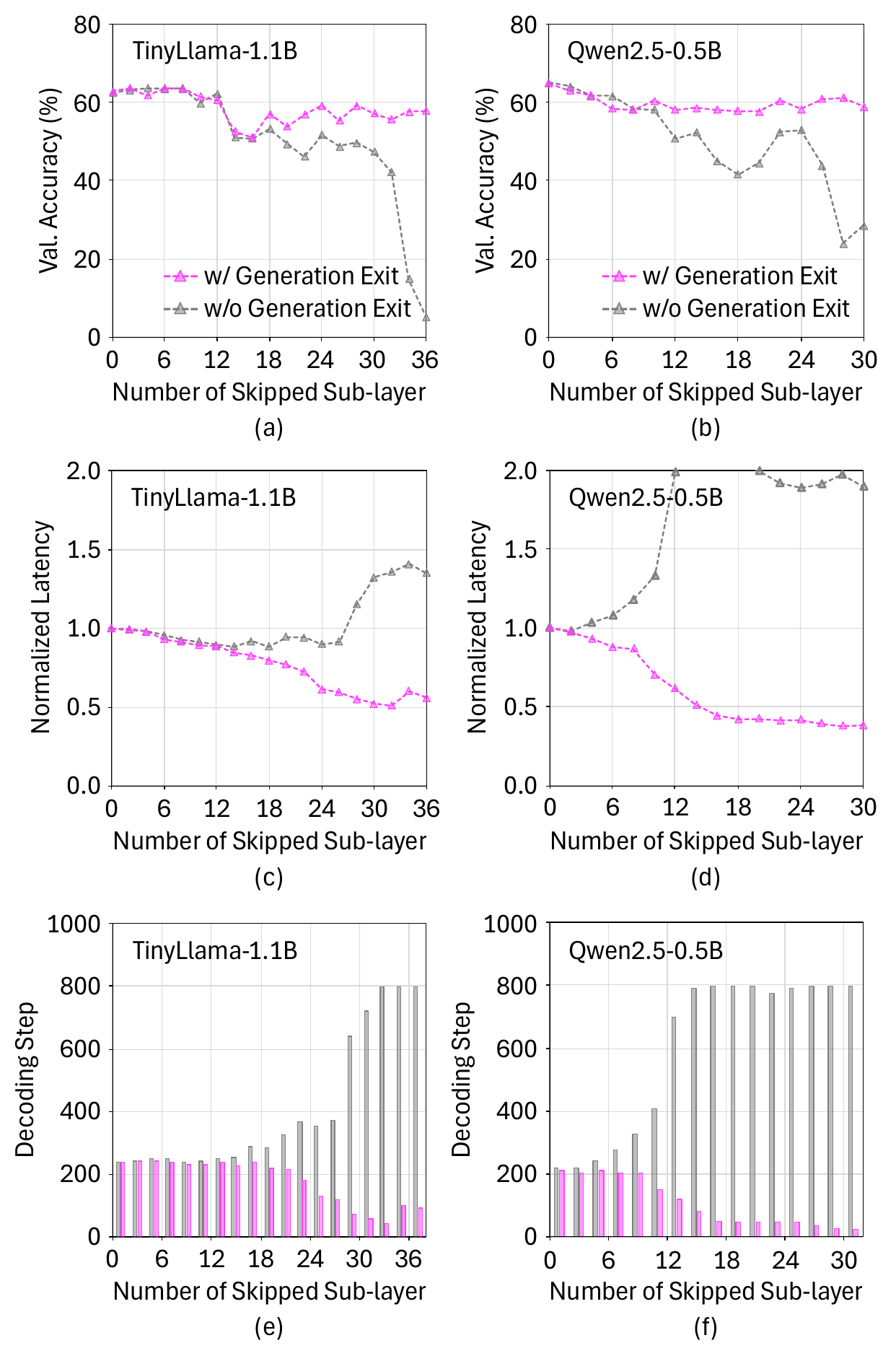}
\vspace{-7.5mm}
\caption{\textbf{Layer Skipping w/ and w/o Generation Exit.} (a)–(b) validation accuracy, (c)–(d) normalized end-to-end latency, and (e)–(f) decoding steps as a function of the number of skipped sub-layers for TinyLlama-1.1B and Qwen2.5-0.5B on the OBQA dataset. Results apply layer skipping only at Stage~2.}
\vspace{-2.5mm}
\label{figure:without_generation_exit}
\end{figure}

\noindent\textbf{With and Without Generation Early Exit.} Figure \ref{figure:without_generation_exit} compares layer skipping with and without generation early exit under the same skip budgets, revealing two regimes. (1) \textit{Mild skipping}. When skipping is small (up to 12 sub-layers for TinyLlama and 2 for Qwen), accuracy, latency, and decoding steps change little, as output confidence remains high and decoding length stays close to the full-layer baseline. In this regime, layer skipping alone provides speedup, so early exit gives limited benefit. (2) \textit{Aggressive skipping}. With larger skips, decoding length grows without early exit, while early exit shortens the sequence and reduces end-to-end latency (Figures~\ref{figure:without_generation_exit}(c)-(f)). The effect appears once decoding steps increase, beyond roughly 16 skipped sub-layers for TinyLlama and 4 for Qwen, indicating that early exit is most beneficial in the high-acceleration regime. Appendix~\ref{app:noexit} further shows that, even without early exit, stage-wise allocation alone outperforms AdaSkip at comparable speedup on the OBQA validation set.

\noindent\textbf{Accuracy Improvements.} Interestingly, in Figures~\ref{figure:without_generation_exit}(a)-(b), the models with generation early exit achieve even higher accuracy than those without it. This indicates that redundantly generated (\textit{i.e.}, low-confidence) tokens under aggressive layer skipping not only increase end-to-end latency but can also harm final predictions. 


\subsection{Evaluation in Other Models} 

\begin{figure}[t]
\centering
\includegraphics[width=\linewidth]{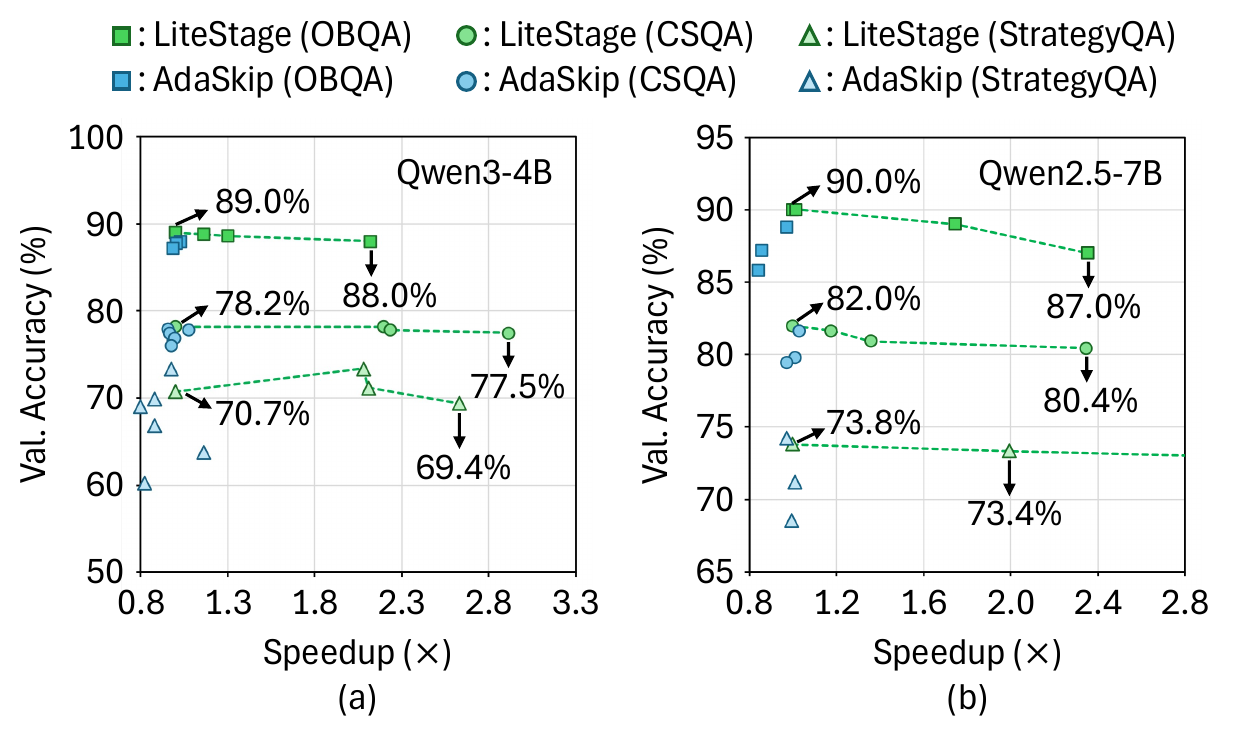}
\vspace{-7.5mm}
\caption{\textbf{Evaluation in 4B and 7B models.} (a) and (b) show the accuracy-speedup trade-off for Qwen3-4B and Qwen2.5-7B, respectively. Speedup is normalized by the end-to-end latency of the full-layer baseline.}
\vspace{-2.5mm}
\label{figure:4b_7b}
\end{figure}

\noindent\textbf{Larger Models (4B/7B).} We further evaluate Qwen3-4B-Instruct-2507 and Qwen2.5-7B-Instruct to verify that our findings generalize beyond small models (${<}$1.1B). As shown in Figure~\ref{figure:4b_7b}, LiteStage consistently outperforms the full-layer baseline and AdaSkip on OBQA, CSQA, and StrategyQA, achieving higher end-to-end speedup while maintaining similar accuracy. The gain is more pronounced in these models, where combining layer skipping with generation early exit yields substantial latency reduction without harming answer quality. Detailed results are provided in Appendix~\ref{sec:supp_4b_7b}.

\noindent\textbf{Deep Reasoning Models.} While LiteStage is designed for short multi-stage reasoning, we also examine whether its strategies extend to \emph{deep reasoning} tasks (\textit{e.g.}, mathematics, coding). To this end, we conduct a diagnostic study on deep reasoning benchmarks, following the similar evaluation protocol used for short multi-stage tasks. In this study, we focus on examining how basic layer skipping behaves in deep reasoning settings and how its performance changes when combined with generation early exit. Please refer to the details in Appendix~\ref{sec:supp_deep_reasoning}.

Our results show that combining layer skipping with generation early exit consistently achieves higher speedups. However, we observe substantial accuracy degradation across \emph{all} evaluated configurations, even under moderate levels of approximation. This behavior indicates that the efficiency-accuracy trade-off in deep reasoning differs fundamentally from that in short multi-stage reasoning. 

%% file: sec/conclusion.tex
\section{Conclusion}
We introduced LiteStage, a latency-aware layer-skipping framework for efficient multi-stage reasoning in small LLMs. By jointly optimizing stage-wise layer budgets and applying confidence-based generation early exit, LiteStage effectively balances accuracy and latency. Experiments on OBQA, CSQA, and StrategyQA demonstrate that LiteStage surpasses prior training-free methods with a large speedup margin. Our results highlight the importance of stage-aware optimization and adaptive decoding for truly efficient multi-stage reasoning.

%% file: sec/appendix.tex
\appendix

\section{Overview}
This appendix provides additional experimental details and analyses that supplement our manuscript "LiteStage: Latency-aware Layer Skipping for Multi-stage Reasoning". We first report statistics on stage-wise decoding length to characterize the computational properties of multi-stage reasoning (Appendix~\ref{sec:supp_data_statistics}), followed by implementation details and training dynamics for decoder-only models adapted from the TinyThinker framework (Appendix~\ref{sec:supp_training_setups}--\ref{sec:supp_training_dynamics}). We then present analyses of sub-layer importance estimation based on cosine similarity, along with extensive ablation studies evaluating the effect of layer skipping across individual reasoning stages, models, and datasets (Appendix~\ref{sec:supp_additional_experimental_results}). These include validation of the similarity proxy and its validation-to-test consistency (Appendix~\ref{sec:supp_importance}), a sensitivity analysis of the early-exit hyperparameters, robustness across evaluation seeds, a comparison with ShortGPT, and stage-wise allocation without generation early exit (Appendix~\ref{app:exit_sweep}--\ref{app:noexit}). We include additional analyses on deep reasoning tasks to examine the behavior of LiteStage under long-horizon generation settings (Appendix~\ref{sec:supp_deep_reasoning}). We further provide additional evaluations on larger models to verify scalability beyond small LMs, as well as experiments under mild distribution shifts to examine the robustness of the validation-selected skip configuration (Appendix~\ref{sec:supp_4b_7b}, Appendix~\ref{sec:supp_dist_shift}).


\section{Data and Implementation Details} 

\subsection{Data Statistics}
\label{sec:supp_data_statistics}
Table~\ref{table:supp_data_statistics} summarizes the average number of decoding steps per reasoning stage in LiteStage. The statistics are estimated using full-layer models, allowing us to analyze the intrinsic complexity of each reasoning stage in its original form, without the influence of layer skipping or early exit. Across all three datasets of OBQA, CSQA, and StrategyQA, the \textit{Analyze} stage consistently exhibits the largest number of decoding steps, indicating that it dominates the overall generation length and computational cost. In contrast, the \textit{Summarize} stage is the shortest, as it typically follows a fixed answer template (\textit{e.g.}, "So, the answer is ($\cdot$)"), requiring minimal generation. The \textit{Recall} stage lies between these extremes, reflecting moderate and relatively stable generation length across datasets. This clear imbalance in stage-wise decoding length motivates our stage-aware optimization strategy, which prioritizes accelerating the longest stages to achieve effective end-to-end speedup.

\begin{table}[t]
\centering
\renewcommand{\arraystretch}{1.1}
\setlength{\tabcolsep}{4pt}
\fontsize{8.5pt}{8.5pt}\selectfont

\caption{\textbf{Decoding Step Statistics.} The number of decoding steps for each stage (Stage~1: Recall, Stage~2: Analyze, and Stage~3: Summarize) is estimated and averaged over the test set of the three datasets (OBQA, CSQA, and StrategyQA) for our models, TinyLlama-1.1B and Qwen2.5-0.5B.}

\begin{tabular}{l|ccc}

\toprule
\textbf{TinyLlama-1.1B} & \textbf{OBQA} & \textbf{CSQA} & \textbf{StrategyQA} \\
\midrule
Recall & 55.5 & 46.4 & 68.6 \\
Analyze & 236.9 & 250.6 & 155.8 \\
Summarize & 7.0 & 7.0 & 7.0 \\

\midrule

\textbf{Qwen2.5-0.5B} & \textbf{OBQA} & \textbf{CSQA} & \textbf{StrategyQA} \\
\midrule
Recall & 49.1 & 36.9 & 58.7 \\
Analyze & 206.4 & 220.1 & 159.8 \\
Summarize & 7.0 & 7.0 & 7.0 \\
\bottomrule
\end{tabular}
\label{table:supp_data_statistics}
\end{table}

\subsection{Training Setups} 
\label{sec:supp_training_setups}

Our training and evaluation pipelines are largely based on the TinyThinker codebase\footnote{https://github.com/shengminp/TinyThinker}. Since the base models such as TinyLlama-1.1B and Qwen2.5-0.5B are not instruction-tuned for multi-stage reasoning, we fine-tune them to follow structured reasoning stages (\textit{e.g.}, recall, analysis, and summarization). The training data consist of questions from the official OBQA, CSQA, and StrategyQA datasets, along with reasoning paths generated by larger models such as GPT, following the TinyThinker protocol. Training on these reasoning paths can therefore be viewed as a form of knowledge distillation from GPT to smaller models such as TinyLlama-1.1B and Qwen2.5-0.5B. Further details of the training procedure can be found in the original TinyThinker paper~\cite{piao2024tinythinker}.

However, TinyThinker’s original experiments are based on T5 models. Accordingly, we adopt a different set of hyperparameters to better support fine-tuning of decoder-only architectures (reported in the main paper). We perform full fine-tuning, updating all model parameters on the complete OBQA, CSQA, and StrategyQA datasets for 10 epochs, with a batch size of 16 on OBQA and CSQA and 24 on StrategyQA, using an initial learning rate of $5\times10^{-5}$. Training uses the AdamW optimizer (default in \texttt{SFTTrainer}) with weight decay 0.0, $\beta_{1}{=}0.9$, $\beta_{2}{=}0.999$, and $\epsilon{=}10^{-8}$. Evaluation follows TinyThinker’s self-consistency (\textit{i.e.}, majority voting) protocol with 10 iterations. All experiments are conducted on a single NVIDIA A6000 GPU (48GB).

\begin{figure*}[t]
\centering
\includegraphics[width=\linewidth]{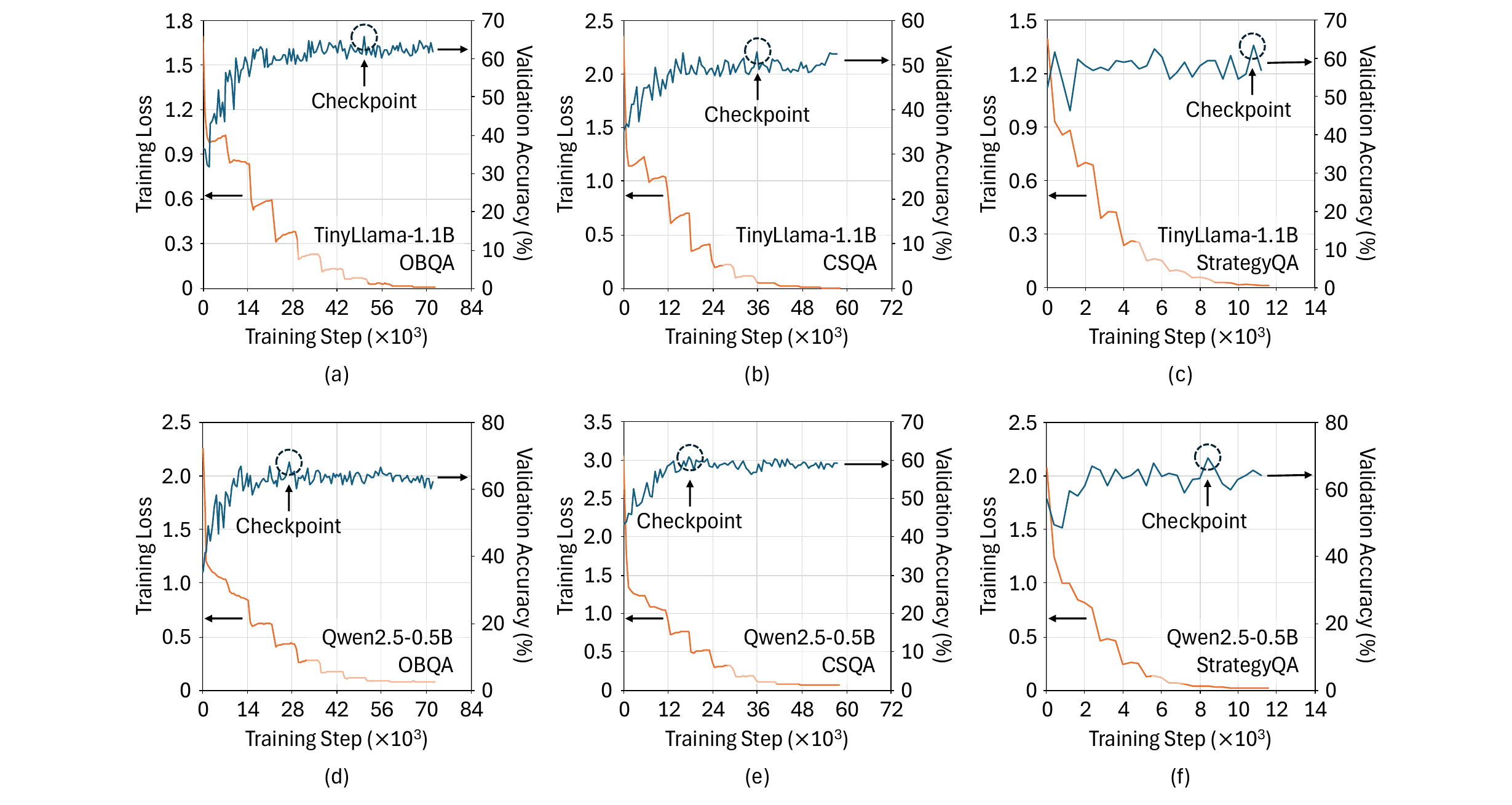}
\caption{\textbf{Training Dynamics.} Training loss and validation accuracy over training steps for TinyLlama-1.1B and Qwen2.5-0.5B on OBQA, CSQA, and StrategyQA. The top row shows results for TinyLlama-1.1B, and the bottom row shows results for Qwen2.5-0.5B. In each plot, training loss is shown on the left y-axis and validation accuracy on the right y-axis. Circles denote the selected checkpoints used for evaluation.}
\label{figure:supp_training}
\end{figure*}

\subsection{Training Dynamics} 
\label{sec:supp_training_dynamics}

Recent progress in reasoning tasks has been largely driven by decoder-only architectures such as LLaMA and Qwen. Motivated by this trend, we extend the TinyThinker framework, that is originally developed for encoder–decoder T5 models, to the LLaMA and Qwen families. Specifically, we fine-tune TinyLlama-1.1B and Qwen2.5-0.5B using TinyThinker’s three-stage reasoning supervision on the OBQA, CSQA, and StrategyQA datasets.

The largest TinyThinker model (T5-Large, 770M parameters) reports test accuracies of 65.4\% (CSQA), 68.8\% (OBQA), and 69.0\% (StrategyQA). In comparison, TinyLlama-1.1B achieves 54.8\%, 64.0\%, and 62.4\% on the same datasets, while Qwen2.5-0.5B attains 64.4\%, 70.2\% , and 65.5\%. Notably, despite having fewer parameters, Qwen2.5-0.5B consistently outperforms TinyLlama-1.1B, reflecting the stronger instruction-following and reasoning priors of the Qwen family. 

Figure~\ref{figure:supp_training} illustrates the training dynamics of both models, including training loss and validation accuracy. Across all datasets, the training loss decreases smoothly, while validation accuracy saturates early and remains stable thereafter. This behavior indicates effective convergence under structured multi-stage reasoning supervision. We also observe that validation accuracy is consistently lower than test accuracy, since test results are computed using self-consistency with ten sampled generations (majority voting), whereas validation accuracy reflects single-pass decoding.

For TinyLlama-1.1B, we attribute part of the remaining performance gap relative to T5-Large to differences in training configuration. TinyThinker’s original experiments utilize four A100 GPUs, enabling substantially larger batch sizes, whereas our experiments are conducted on a single A6000 GPU. This difference likely affects optimization stability and final performance, particularly for models trained to follow multi-stage reasoning patterns.

\begin{figure*}[t]
\centering
\includegraphics[width=\linewidth]{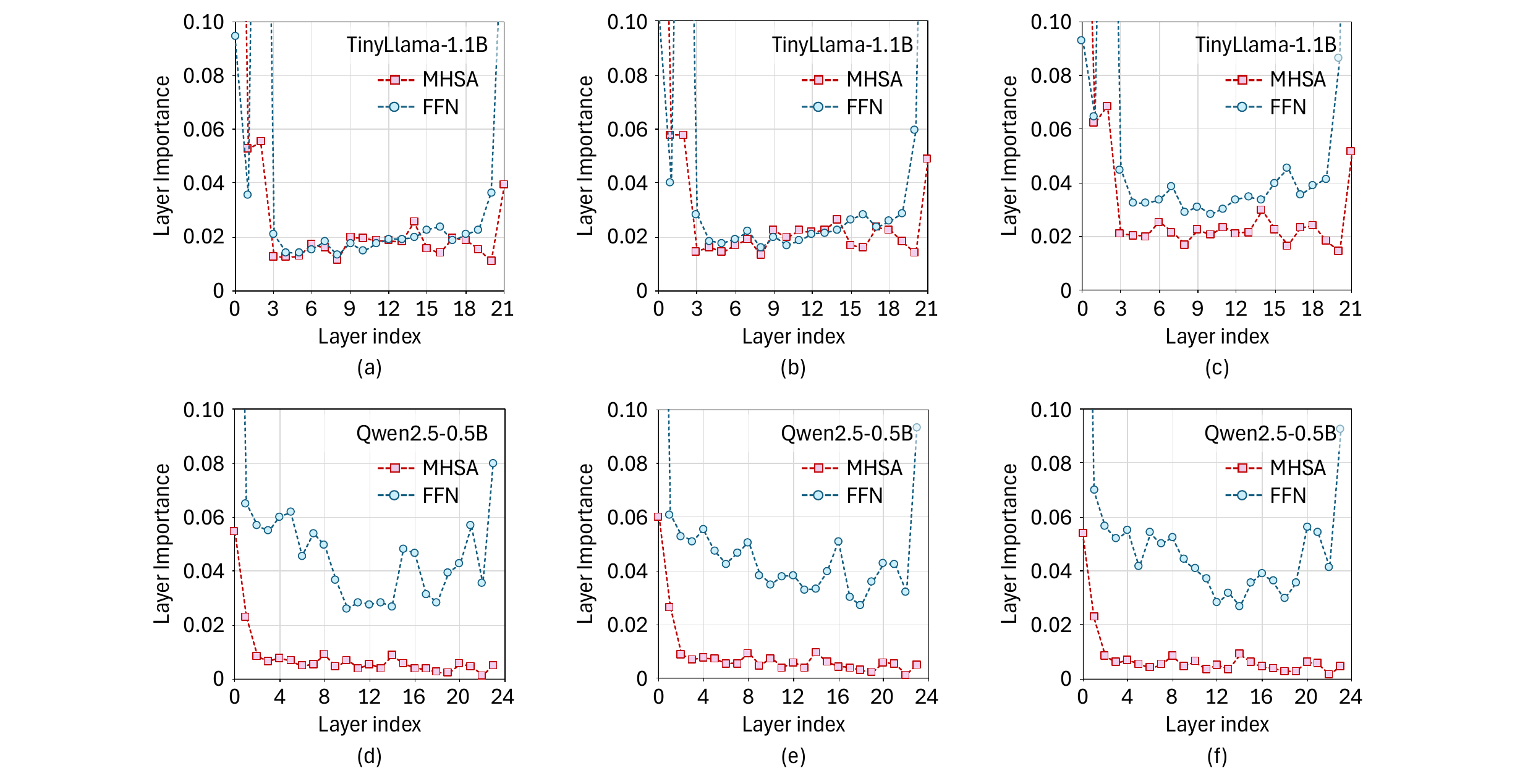}
\caption{\textbf{Layer Importance.} Sub-layer-wise importance estimates for TinyLlama-1.1B (a)–(c) and Qwen2.5-0.5B (d)–(f) across OBQA, CSQA, and StrategyQA. Importance is computed using cosine similarity between sub-layer inputs and outputs, separately for multi-head self-attention (MHSA) and feed-forward network (FFN) layers.}
\label{figure:supp_importance}
\end{figure*}

\section{Additional Experimental Results}
\label{sec:supp_additional_experimental_results}

\subsection{Example of Generation Early Exit}
\begin{figure*}[t]
\centering
\includegraphics[width=\linewidth]{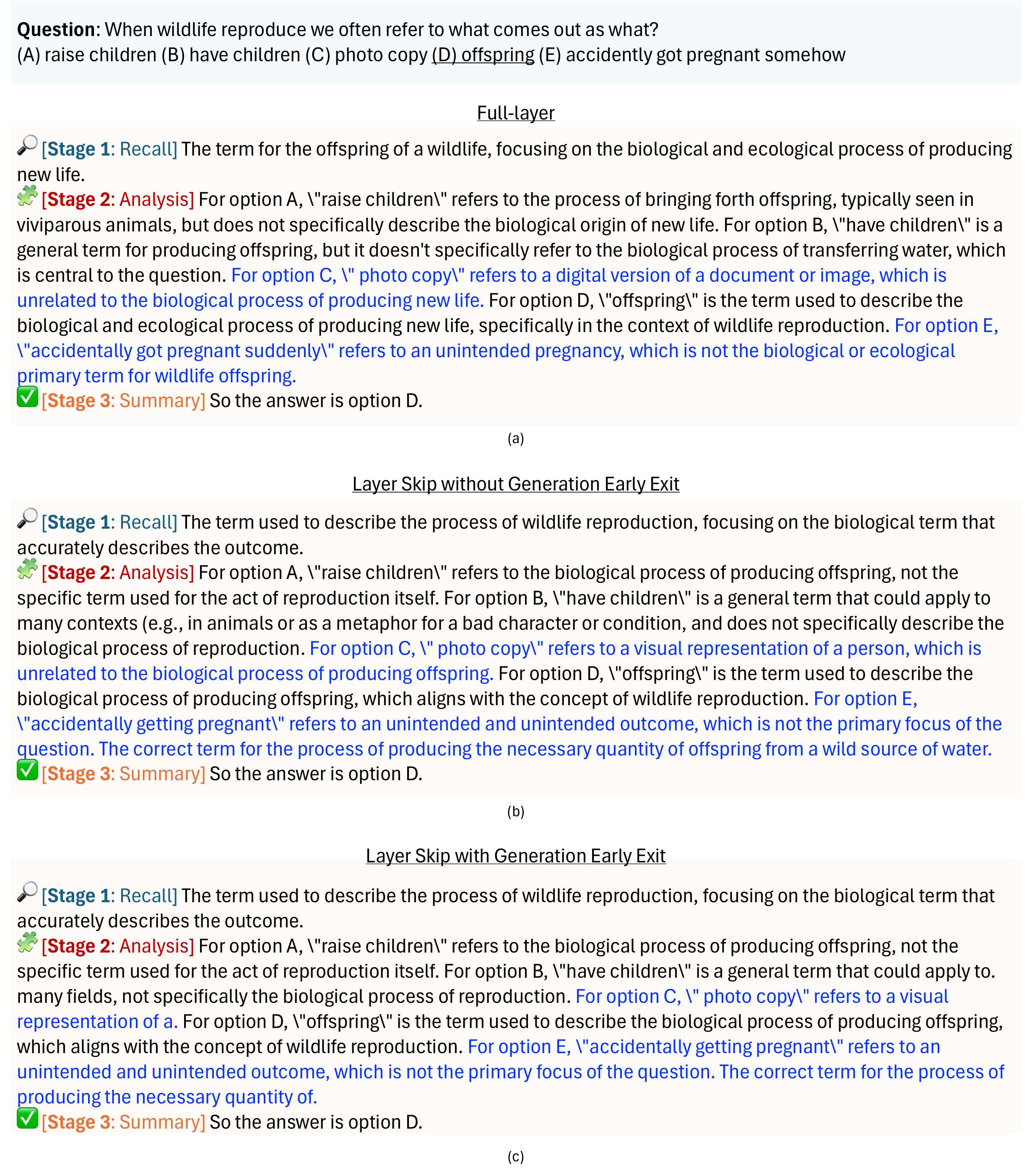}
\caption{\textbf{Example of Generation Early Exit.} (a)-(c) the three-stage reasoning outcomes from models with full-layer, with layer skipping but without generation early exit, and with layer skipping and generation early eixt in a CSQA test sample.}
\label{figure:supp_generation_exit_example}
\end{figure*}

Figure~\ref{figure:supp_generation_exit_example} presents a CSQA test example comparing three-stage reasoning outputs produced by models with and without generation early exit. Due to the size of the visualization, the figure is placed on later pages. For reference, the full-layer baseline output is shown in Figure~\ref{figure:supp_generation_exit_example}(a), while Figures~\ref{figure:supp_generation_exit_example}(b) and (c) correspond to layer skipping without and with generation early exit, respectively.

Generation early exit terminates decoding once token-level confidence falls below a predefined threshold. As a result, the generated tokens remain identical to the baseline up to the exit point, and only the low-confidence suffix is truncated. In the figure, the truncated segments are highlighted in blue for clarity. For example, the full-layer reasoning phrase

\begin{quote}
\small
\texttt{For option C, photo copy refers to a visual representation of a person, which is unrelated to the biological process of producing offspring,}
\end{quote}

is shortened to

\begin{quote}
\small
\texttt{For option C, photo copy refers to a visual representation of a}
\end{quote}

This behavior indicates that the model maintains high confidence up to the phrase "visual representation", while the subsequent explanatory tokens exhibit lower confidence and contribute marginally to decision making. Importantly, despite the truncated reasoning, both models arrive at the same final conclusion in Stage~3, correctly rejecting option~C and selecting option~D as the answer. This example demonstrates that generation early exit effectively removes redundant low-confidence reasoning tokens without altering the final prediction, thereby reducing decoding length while preserving reasoning correctness.

\subsection{Layer Importance Estimation}
\label{sec:supp_importance} 
Figure \ref{figure:supp_importance} presents the sub-layer-wise importance estimates obtained in Step 1 of our pipeline. It reveals several consistent patterns in sub-layer importance that directly inform our skipping strategy. Across all three datasets, the relative ordering of layer importance remains largely stable, suggesting that the contribution of individual layers is not highly task-specific. However, we observe dataset-dependent shifts in the balance between multi-head self-attention (MHSA) and feed-forward network (FFN) sub-layers.

In particular, for TinyLlama-1.1B, FFN sub-layers on StrategyQA exhibit comparatively higher importance than on OBQA and CSQA, leading the resulting skipping policy to prioritize MHSA layers while preserving FFN computation. A more pronounced distinction emerges across model families. In Qwen2.5-0.5B, MHSA sub-layers consistently demonstrate substantially higher importance than FFN sub-layers across all datasets. Consequently, layer skipping in Qwen models is predominantly applied to MHSA, whereas FFN layers are largely retained.

Despite these differences in absolute importance levels, the overall layer-wise trends remain consistent across datasets, indicating that sub-layer importance is primarily governed by model architecture rather than dataset-specific characteristics. These observations motivate our use of sub-layer–level importance estimation and justify decoupling the skipping decisions for MHSA and FFN, enabling LiteStage to adapt its computation budget to both architectural and dataset-level variations.

\paragraph{Validation--test consistency.}
To verify that importance estimated on validation data transfers to test data,
we profile sub-layer cosine similarity separately on each split and compare the
rankings (Table~\ref{tab:valtest}). Rankings are nearly identical
(Spearman $\rho \geq 0.994$), and the most-skippable sub-layers largely
coincide; the few mismatches are single-position swaps.

\begin{table}[H]
\centering
\small
\caption{Consistency of sub-layer importance rankings between validation and
test sets (44 candidates for TinyLlama, 48 for Qwen; MHSA and FFN combined).}
\label{tab:valtest}
\resizebox{\linewidth}{!}{%
\begin{tabular}{llcccc}
\toprule
Model & Dataset & $\rho$ & Top-4 & Top-8 & Top-12 \\
\midrule
\multirow{3}{*}{TinyLlama-1.1B} & OBQA & 0.999 & 4/4 & 8/8 & 12/12 \\
 & CSQA & 1.000 & 4/4 & 8/8 & 12/12 \\
 & StrategyQA & 0.999 & 4/4 & 8/8 & 12/12 \\
\midrule
\multirow{3}{*}{Qwen2.5-0.5B} & OBQA & 1.000 & 4/4 & 7/8 & 12/12 \\
 & CSQA & 1.000 & 4/4 & 8/8 & 12/12 \\
 & StrategyQA & 0.994 & 3/4 & 6/8 & 12/12 \\
\bottomrule
\end{tabular}}
\end{table}

\paragraph{Validity of the similarity proxy.}
We rank all candidate sub-layers of TinyLlama-1.1B on OBQA by cosine
similarity, split them into quartiles, and skip four sub-layers from each
quartile (Table~\ref{tab:quartile}). Accuracy decreases monotonically from Q1
to Q3 and drops sharply only at Q4 (lowest similarity), indicating that cosine
similarity reliably identifies the small set of critical sub-layers.

\begin{table}[H]
\centering
\small
\caption{Skipping four sub-layers from each cosine-similarity quartile
(TinyLlama-1.1B, OBQA validation, mean $\pm$ std over three seeds).}
\label{tab:quartile}
\begin{tabular}{llc}
\toprule
Quartile & Similarity tier & Acc (\%) \\
\midrule
Q1 & top 25\% (high) & 63.07 $\pm$ 1.21 \\
Q2 & 25--50\% & 62.53 $\pm$ 0.70 \\
Q3 & 50--75\% & 61.07 $\pm$ 0.90 \\
Q4 & bottom 25\% (low) & 26.07 $\pm$ 1.01 \\
\bottomrule
\end{tabular}
\end{table}


\subsection{Ablation Studies}

\begin{figure*}[t]
\centering
\includegraphics[width=\linewidth]{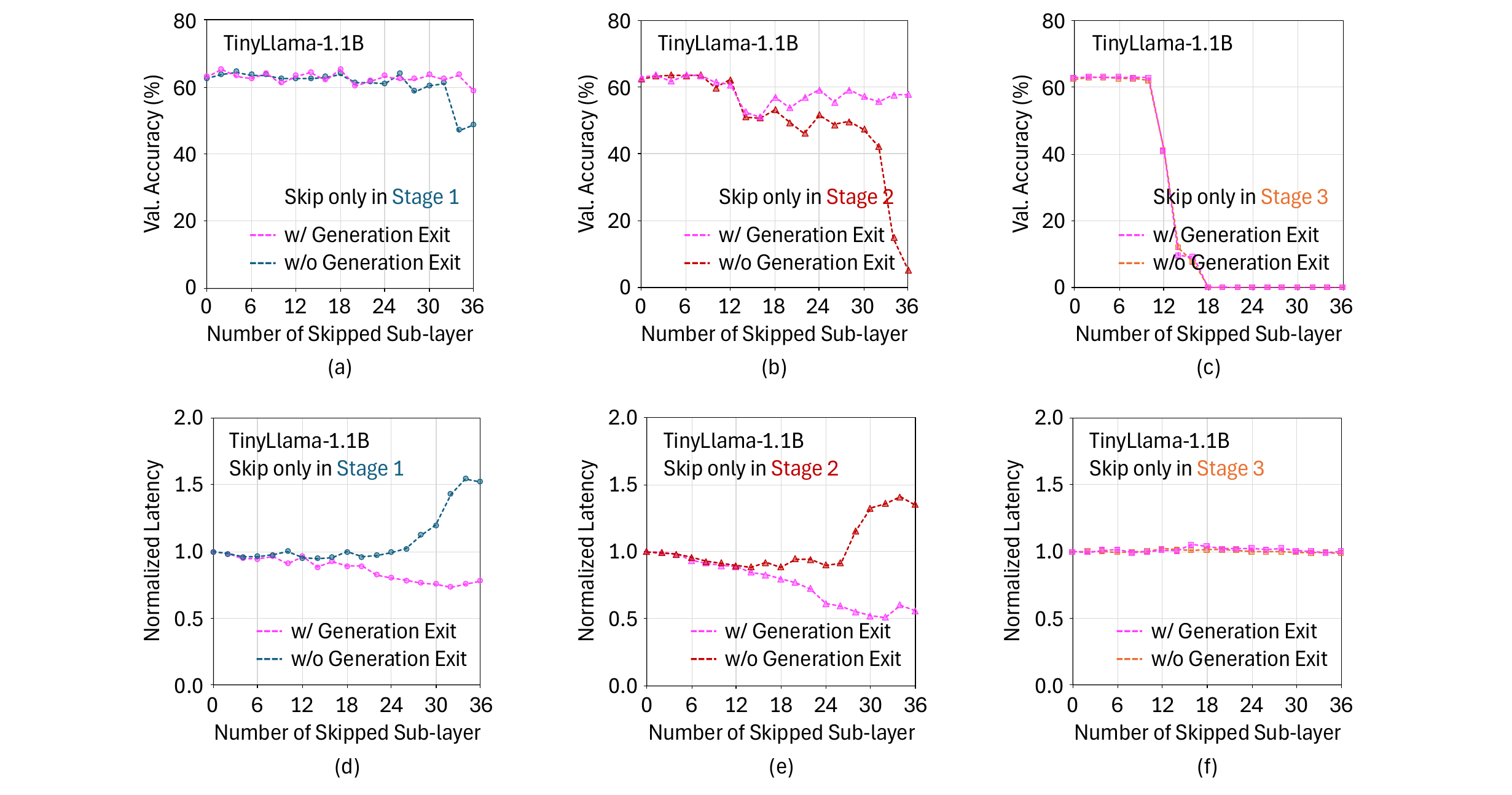}
\caption{\textbf{Ablation: TinyLlama-1.1B on OBQA.} Validation accuracy (top row) and normalized latency (bottom row) as a function of the number of skipped sub-layers when applying layer skipping to a single reasoning stage at a time (Stage~1, Stage~2, or Stage~3). Results are shown for TinyLlama-1.1B on OBQA, comparing configurations with and without generation early exit.}
\label{figure:supp_ablation_tinyllama_obqa}
\end{figure*}

\begin{figure*}[t]
\centering
\includegraphics[width=\linewidth]{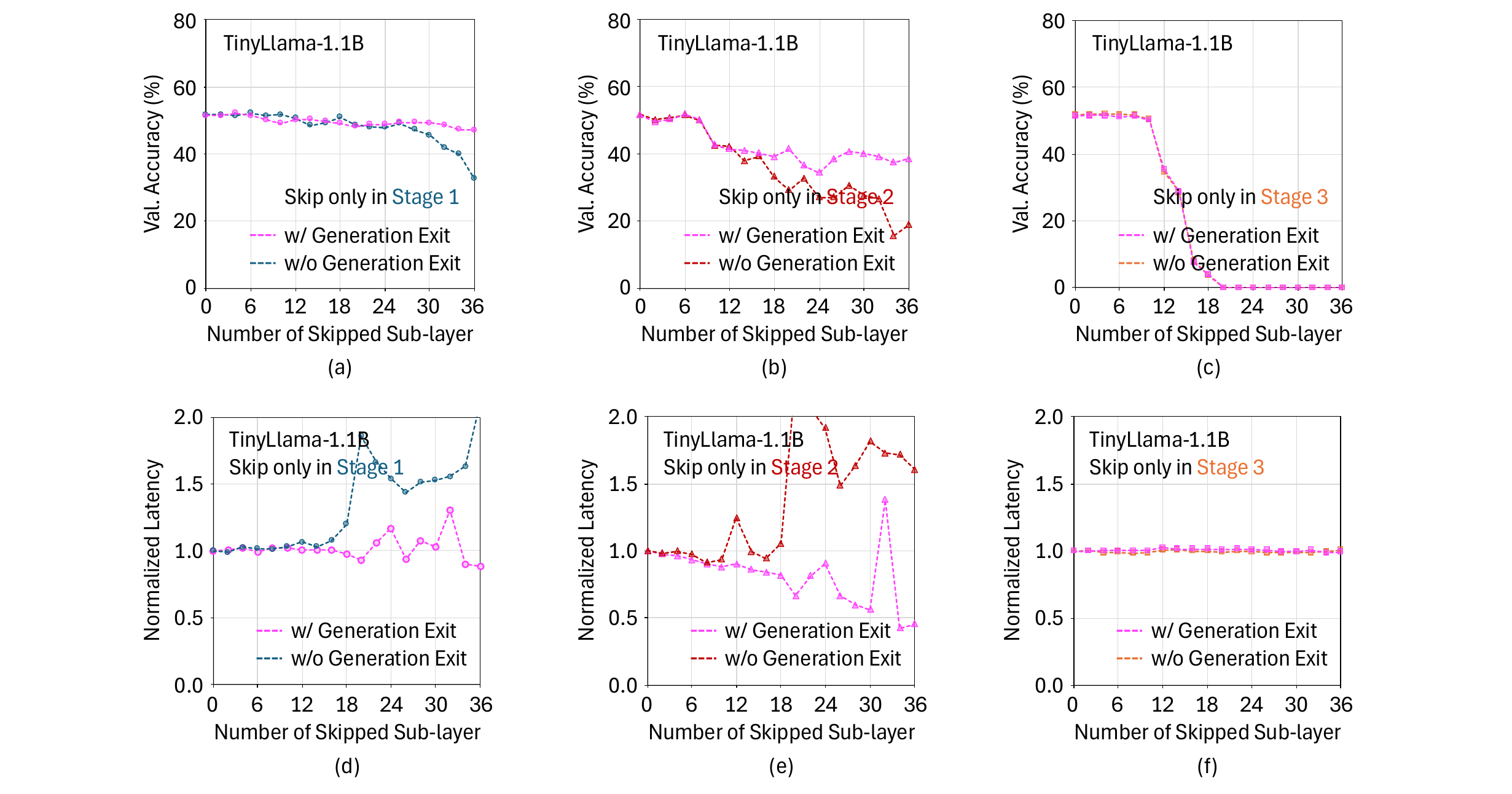}
\caption{\textbf{Ablation: TinyLlama-1.1B on CSQA.} Validation accuracy (top row) and normalized latency (bottom row) as a function of the number of skipped sub-layers when applying layer skipping to a single reasoning stage at a time (Stage~1, Stage~2, or Stage~3). Results are shown for TinyLlama-1.1B on CSQA, comparing configurations with and without generation early exit.}
\label{figure:supp_ablation_tinyllama_csqa}
\end{figure*}

\begin{figure*}[t]
\centering
\includegraphics[width=\linewidth]{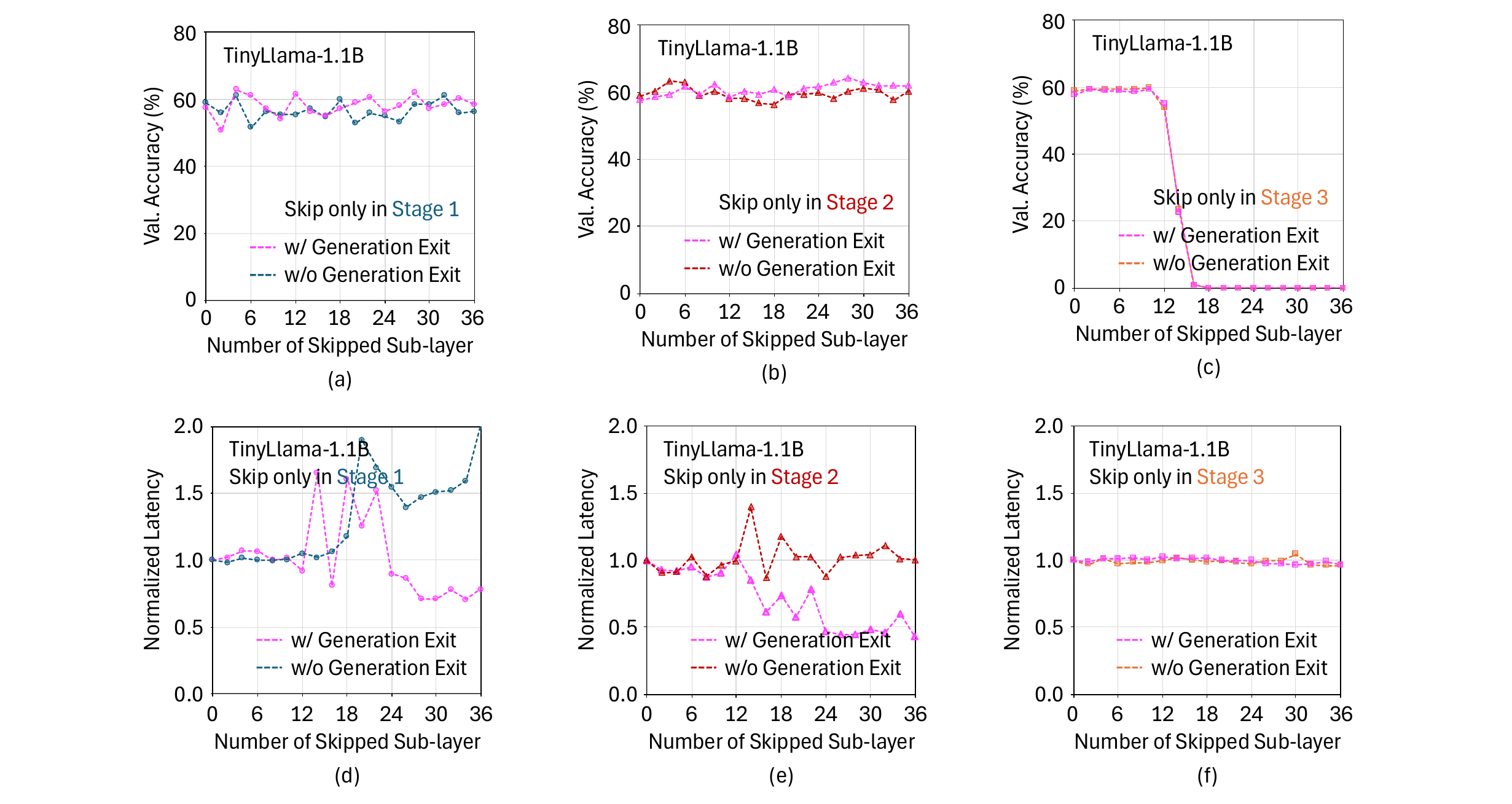}
\caption{\textbf{Ablation: TinyLlama-1.1B on StrategyQA.} Validation accuracy (top row) and normalized latency (bottom row) as a function of the number of skipped sub-layers when applying layer skipping to a single reasoning stage at a time (Stage~1, Stage~2, or Stage~3). Results are shown for TinyLlama-1.1B on StrategyQA, comparing configurations with and without generation early exit.}
\label{figure:supp_ablation_tinyllama_strategyqa}
\end{figure*}

\begin{figure*}[t]
\centering
\includegraphics[width=\linewidth]{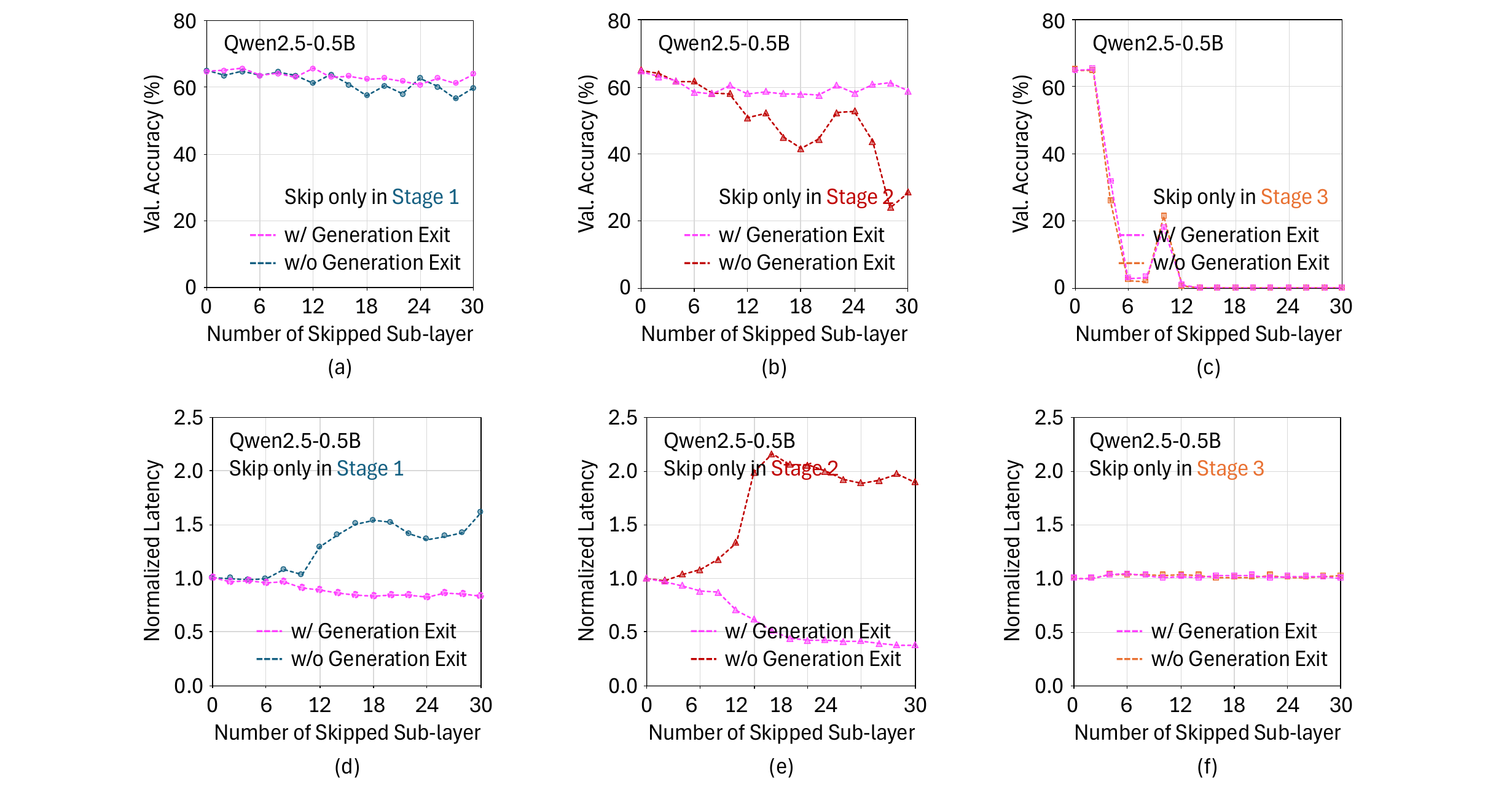}
\caption{\textbf{Ablation: Qwen2.5-0.5B on OBQA.} Validation accuracy (top row) and normalized latency (bottom row) as a function of the number of skipped sub-layers when applying layer skipping to a single reasoning stage at a time (Stage~1, Stage~2, or Stage~3). Results are shown for Qwen2.5-0.5B on OBQA, comparing configurations with and without generation early exit.}
\label{figure:supp_ablation_qwen_obqa}
\end{figure*}

\begin{figure*}[t]
\centering
\includegraphics[width=\linewidth]{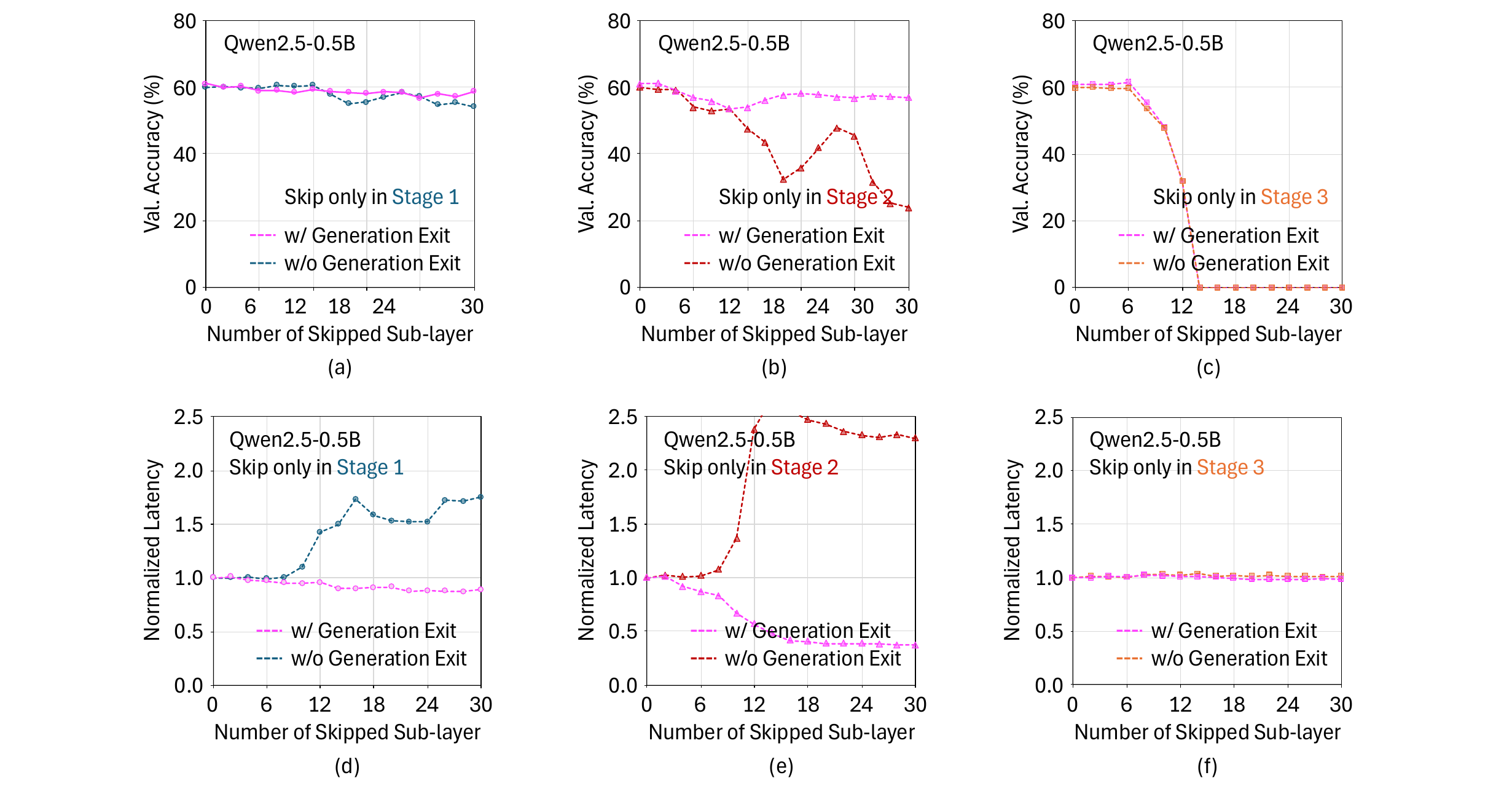}
\caption{\textbf{Ablation: Qwen2.5-0.5B on CSQA.} Validation accuracy (top row) and normalized latency (bottom row) as a function of the number of skipped sub-layers when applying layer skipping to a single reasoning stage at a time (Stage~1, Stage~2, or Stage~3). Results are shown for Qwen2.5-0.5B on CSQA, comparing configurations with and without generation early exit.}
\label{figure:supp_ablation_qwen_csqa}
\end{figure*}

\begin{figure*}[t]
\centering
\includegraphics[width=\linewidth]{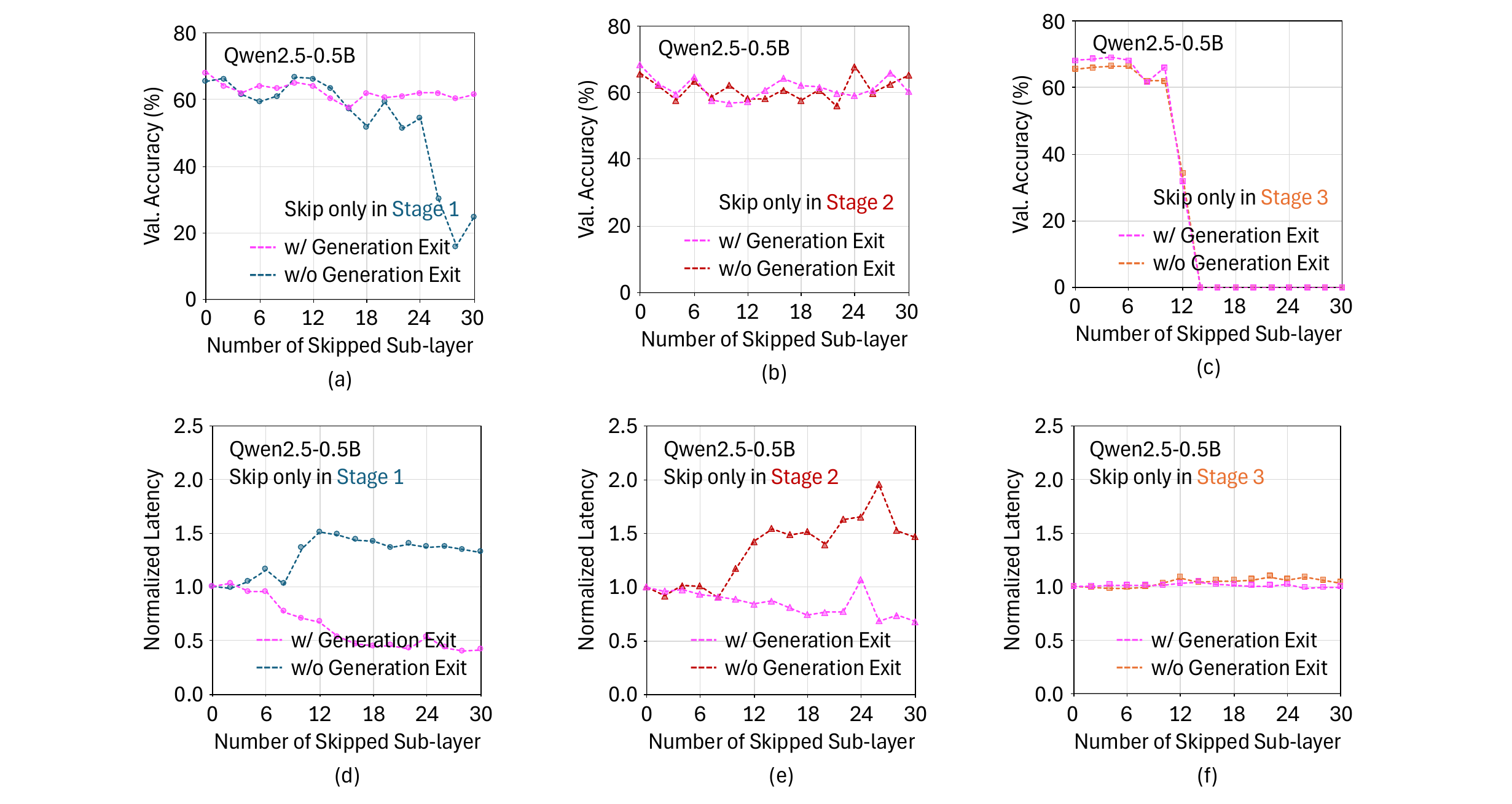}
\caption{\textbf{Ablation: Qwen2.5-0.5B on StrategyQA.} Validation accuracy (top row) and normalized latency (bottom row) as a function of the number of skipped sub-layers when applying layer skipping to a single reasoning stage at a time (Stage~1, Stage~2, or Stage~3). Results are shown for Qwen2.5-0.5B on StrategyQA, comparing configurations with and without generation early exit.}
\label{figure:supp_ablation_qwen_strategyqa}
\end{figure*}

To analyze stage-wise accuracy and latency behavior under layer skipping, we conduct extensive ablation studies on two models, TinyLlama-1.1B and Qwen2.5-0.5B, across three benchmarks: OBQA, CSQA, and StrategyQA. In these experiments, we vary the number of skipped sub-layers while applying layer skipping to a single reasoning stage at a time (Stage 1, Stage 2, or Stage 3), and measure validation accuracy along with normalized end-to-end latency, where 1.0 corresponds to the full-layer baseline. We further ablate each configuration with and without generation early exit to isolate its effect when combined with layer skipping. Due to space constraints, all ablation figures are provided in the later pages: Figures~\ref{figure:supp_ablation_tinyllama_obqa}-\ref{figure:supp_ablation_tinyllama_strategyqa} report results for TinyLlama-1.1B, and Figures~\ref{figure:supp_ablation_qwen_obqa}-\ref{figure:supp_ablation_qwen_strategyqa} report results for Qwen2.5-0.5B.

From these ablations, we highlight two key observations: (1) the importance of explicitly profiling the accuracy–latency trade-off, and (2) the consistent benefits of generation early exit across models, datasets, and reasoning stages.

(1) \textit{importance of accuracy-latency profiling:} Across all datasets, we observe a broadly consistent sensitivity pattern across reasoning stages: Stage~1 is generally the most robust to layer skipping, followed by Stage~2, while Stage~3 is the most sensitive. However, a critical finding is that the degree of sensitivity varies substantially across both models and datasets. For example, when skipping only Stage~2 on OBQA, Qwen2.5-0.5B exhibits markedly higher accuracy robustness than TinyLlama-1.1B, as shown in Figures~\ref{figure:supp_ablation_tinyllama_obqa}(b) and ~\ref{figure:supp_ablation_qwen_obqa}(b). Such differences are non-trivial and cannot be reliably inferred from static proxy metrics such as cosine similarity alone; instead, they only become apparent through direct profiling of the accuracy-latency behavior.

Moreover, the ablations reveal that accuracy and latency do not vary monotonically with the number of skipped sub-layers. In some cases, skipping more layers unexpectedly yields higher accuracy, or skipping fewer layers results in higher speedup. For instance, in Figure~\ref{figure:supp_ablation_tinyllama_obqa}(a), skipping 18 sub-layers achieves higher validation accuracy than skipping 16 sub-layers, while in Figure~\ref{figure:supp_ablation_tinyllama_obqa}(d), skipping 16 sub-layers results in higher normalized latency than skipping 14 sub-layers, even when generation early exit is applied. These irregularities highlight that naïvely increasing or decreasing the skip budget can lead to suboptimal or even counterproductive outcomes. 

LiteStage’s "Step 2: Search Layer Budget" directly addresses this challenge by selecting the optimal number of skipped sub-layers based on the empirically observed accuracy-latency profile, rather than relying on monotonic assumptions or heuristic thresholds. This search-based strategy allows LiteStage to avoid suboptimal configurations that would otherwise degrade end-to-end performance.

(2) \textit{benefits of generation early exit:} A second consistent observation across all ablations is the crucial role of generation early exit in realizing practical latency gains. Without early exit, aggressive layer skipping often increases generation length, which can negate or even reverse latency improvements. This phenomenon is clearly visible across all datasets and both models, particularly when skipping Stage 2, where generation length is typically the longest.

By incorporating generation early exit, latency growth is effectively suppressed, and normalized latency remains stable or decreases even under aggressive skipping. This trend is consistently observed across OBQA, CSQA, and StrategyQA for both TinyLlama-1.1B and Qwen2.5-0.5B. While Figure~\ref{figure:without_generation_exit} (main paper) already demonstrates this effect on a representative setting, the supplementary ablations confirm that the benefit of generation early exit generalizes across models, datasets, and reasoning stages.

Taken together, these results underscore that layer skipping alone is insufficient for achieving reliable end-to-end acceleration. Instead, effective efficiency gains require joint consideration of stage-wise accuracy-latency profiling and generation early exit, motivating their integrated use in LiteStage.

\begin{table*}[t]
\centering
\renewcommand{\arraystretch}{1.15}
\setlength{\tabcolsep}{4pt}
\fontsize{8.5pt}{8.5pt}\selectfont

\caption {\textbf{Deep Reasoning Performance.} Test accuracy, end-to-end speedup, and the number of decoding steps are estimated on the three benchmarks, AIME 2025 (AIME 25), GPQA-Diamond (GPQA-D), and LiveCodeBench v5 (LCB-v5). The speedup is normalized by the full-layer baseline, and the decoding steps are averaged over test samples. LS, PF, and GE denote Layer Skipping, Periodic Full-decoding, and Generation Early Exit, respectively. Results are experimented using Qwen3-1.7B.}


\begin{tabular}{l|c|ccc|ccc|ccc}

\toprule
\multirow{2}{*}{\textbf{Method}} & \multirow{2}{*}{\textbf{\# Skip}} & \multicolumn{3}{c}{\textbf{Accuracy (Pass@1)}} & \multicolumn{3}{c}{\textbf{End-to-End Speedup ($\times$)}} & \multicolumn{3}{c}{\textbf{Decoding Steps}} \\ 
& & AIME 25 & GPQA-D & LCB-v5 & AIME 25 & GPQA-D & LCB-v5 & AIME 25 & GPQA-D & LCB-v5 \\ 

\midrule

Full-layer & 0 & 30.0 & 46.0 & 34.8 & 1.00 & 1.00 & 1.00 & 18530 & 9374 & 16175 \\
\midrule
LS & 1 & 30.0 & 32.8 & 34.1 & 0.91 & 0.75 & 0.88 & 22230 & 12511 & 19989 \\
& 2 & 13.3 & 30.8 & 11.8 & 1.05 & 1.48 & 0.98 & 20584 & 7195 & 18655 \\
& 3 &  0.0 & 20.2 &  0.0 & 0.86 & 0.92 & 0.68 & 25681 & 11706 & 27273 \\

\midrule

LS+PF& 1 & 26.7 & 38.9 & 34.8 & 0.91 & 0.89 & 0.93 & 20221 & 10523 & 18108 \\
& 2 & 20.0 & 35.9 & 35.5 & 1.06 & 1.10 & 0.96 & 18992 & 8766 & 17224 \\
& 3 & 20.0 & 32.8 & 25.5 & 1.11 & 1.15 & 1.04 & 18149 & 8714 & 16780 \\

\midrule

LS+PF+GE& 1 & 26.7 & 36.4 & 31.2 & 1.06 & 1.08 & 1.35 & 18681 & 8808 & 12568 \\
& 2 & 20.0 & 33.3 & 26.2 & 1.40 & 1.84 & 1.36 & 14699 & 5296 & 12643 \\
& 3 & 16.7 & 31.8 & 17.6 & 1.28 & 1.36 & 1.46 & 15768 & 7081 & 11863 \\

\bottomrule

\end{tabular}
\label{table:supp_deep_reasoning}
\end{table*}

\subsection{Sensitivity to Early-Exit Hyperparameters}
\label{app:exit_sweep}

We sweep the confidence window $n \in \{3,5,7,10\}$ and threshold
$\tau \in \{0.4,0.5,0.6\}$ on the OBQA validation set (Table~\ref{tab:exit_sweep}).
On TinyLlama-1.1B, the default ($n{=}5$, $\tau{=}0.5$) attains the highest
accuracy (60.8\%), and accuracy varies within 3.2\%p across the grid.
On Qwen2.5-0.5B, the default is not the strict optimum (62.0\% vs.\ 64.0\% at
$n{=}3$, $\tau{=}0.6$) but lies on a wide plateau (58.8--64.0\%); applying
Qwen's optimum to TinyLlama lowers accuracy (60.8\%$\rightarrow$58.4\%).
We therefore use the same default in all settings without per-setting tuning.

\begin{table}[H]
\centering
\small
\caption{Early-exit hyperparameter sweep on OBQA (validation). Latency is
total end-to-end latency (sec). Skip budgets (Stage 1/2/3): TinyLlama 38/12/8,
Qwen 22/4/2 sub-layers. Default setting in bold.}
\label{tab:exit_sweep}
\begin{tabular}{cc cc cc}
\toprule
 & & \multicolumn{2}{c}{TinyLlama-1.1B} & \multicolumn{2}{c}{Qwen2.5-0.5B} \\
\cmidrule(lr){3-4}\cmidrule(lr){5-6}
$n$ & $\tau$ & Acc (\%) & Latency & Acc (\%) & Latency \\
\midrule
3  & 0.4 & 59.0 & 117.5 & 60.4 & 101.9 \\
5  & 0.4 & 58.0 & 137.6 & 61.6 & 115.0 \\
7  & 0.4 & 57.6 & 147.6 & 62.8 & 117.9 \\
10 & 0.4 & 59.8 & 150.1 & 59.6 & 126.6 \\
\midrule
3  & 0.5 & 60.0 & 111.9 & 61.6 & 85.1 \\
\textbf{5} & \textbf{0.5} & \textbf{60.8} & \textbf{116.4} & \textbf{62.0} & \textbf{99.2} \\
7  & 0.5 & 60.6 & 130.6 & 61.6 & 106.6 \\
10 & 0.5 & 58.8 & 155.5 & 58.8 & 115.8 \\
\midrule
3  & 0.6 & 58.4 & 102.2 & 64.0 & 62.2 \\
5  & 0.6 & 59.2 & 107.9 & 62.0 & 79.3 \\
7  & 0.6 & 60.0 & 125.3 & 62.8 & 88.8 \\
10 & 0.6 & 60.4 & 136.9 & 62.2 & 103.3 \\
\bottomrule
\end{tabular}
\end{table}

\subsection{Robustness across Evaluation Seeds}
\label{app:multiseed}

We re-evaluate representative operating points with three evaluation seeds
(different sampling seeds for self-consistency voting; the trained checkpoint
is fixed). Table~\ref{tab:multiseed} compares LiteStage and AdaSkip in the
\emph{comparable} region (where the two are closest) and the \emph{aggressive}
region. The standard deviation is small ($\leq$1.5\%p at most points), and in
the aggressive region the gap far exceeds seed variance. Note that re-training
with multiple initialization seeds was not performed (see Limitations).

\begin{table*}[t]
\centering
\small
\caption{Mean $\pm$ std over three evaluation seeds. Numbers in parentheses
denote AdaSkip's skipped sub-layers.}
\label{tab:multiseed}
\begin{tabular}{llllcc}
\toprule
Model & Dataset & Region & Method & Acc (\%) & Speedup ($\times$) \\
\midrule
\multirow{12}{*}{TinyLlama-1.1B}
 & \multirow{4}{*}{OBQA} & \multirow{2}{*}{Compar.} & AdaSkip (8) & 62.33 $\pm$ 0.76 & 1.112 $\pm$ 0.013 \\
 & & & LiteStage & 63.07 $\pm$ 0.42 & 1.108 $\pm$ 0.013 \\
 & & \multirow{2}{*}{Aggr.} & AdaSkip (14) & 2.80 $\pm$ 0.53 & 1.143 $\pm$ 0.007 \\
 & & & LiteStage & 59.87 $\pm$ 0.70 & 1.517 $\pm$ 0.027 \\
\cmidrule(lr){2-6}
 & \multirow{4}{*}{CSQA} & \multirow{2}{*}{Compar.} & AdaSkip (4) & 52.88 $\pm$ 0.59 & 1.082 $\pm$ 0.010 \\
 & & & LiteStage & 53.34 $\pm$ 0.93 & 1.094 $\pm$ 0.006 \\
 & & \multirow{2}{*}{Aggr.} & AdaSkip (10) & 45.48 $\pm$ 0.13 & 1.107 $\pm$ 0.011 \\
 & & & LiteStage & 52.99 $\pm$ 0.57 & 1.106 $\pm$ 0.005 \\
\cmidrule(lr){2-6}
 & \multirow{4}{*}{StrategyQA} & \multirow{2}{*}{Compar.} & AdaSkip (2) & 59.83 $\pm$ 3.73 & 1.017 $\pm$ 0.075 \\
 & & & LiteStage & 59.53 $\pm$ 0.67 & 1.667 $\pm$ 0.097 \\
 & & \multirow{2}{*}{Aggr.} & AdaSkip (18) & 0.00 $\pm$ 0.00 & 0.599 $\pm$ 0.011 \\
 & & & LiteStage & 58.51 $\pm$ 0.76 & 1.959 $\pm$ 0.054 \\
\midrule
\multirow{12}{*}{Qwen2.5-0.5B}
 & \multirow{4}{*}{OBQA} & \multirow{2}{*}{Compar.} & AdaSkip (2) & 66.47 $\pm$ 1.55 & 1.006 $\pm$ 0.011 \\
 & & & LiteStage & 66.00 $\pm$ 0.53 & 1.132 $\pm$ 0.024 \\
 & & \multirow{2}{*}{Aggr.} & AdaSkip (4) & 30.13 $\pm$ 0.31 & 0.867 $\pm$ 0.018 \\
 & & & LiteStage & 62.67 $\pm$ 0.76 & 1.328 $\pm$ 0.029 \\
\cmidrule(lr){2-6}
 & \multirow{4}{*}{CSQA} & \multirow{2}{*}{Compar.} & AdaSkip (4) & 60.00 $\pm$ 0.40 & 1.011 $\pm$ 0.009 \\
 & & & LiteStage & 60.71 $\pm$ 0.17 & 1.200 $\pm$ 0.103 \\
 & & \multirow{2}{*}{Aggr.} & AdaSkip (10) & 34.10 $\pm$ 0.59 & 0.669 $\pm$ 0.004 \\
 & & & LiteStage & 60.39 $\pm$ 0.37 & 1.293 $\pm$ 0.122 \\
\cmidrule(lr){2-6}
 & \multirow{4}{*}{StrategyQA} & \multirow{2}{*}{Compar.} & AdaSkip (4) & 62.01 $\pm$ 0.44 & 1.071 $\pm$ 0.023 \\
 & & & LiteStage & 62.59 $\pm$ 0.50 & 1.373 $\pm$ 0.021 \\
 & & \multirow{2}{*}{Aggr.} & AdaSkip (8) & 52.55 $\pm$ 1.40 & 0.899 $\pm$ 0.052 \\
 & & & LiteStage & 60.41 $\pm$ 1.40 & 1.637 $\pm$ 0.114 \\
\bottomrule
\end{tabular}
\end{table*}

\subsection{Comparison with ShortGPT}
\label{app:shortgpt}

We additionally compare with ShortGPT~\citep{men2024shortgpt}, a training-free
block-pruning method (Table~\ref{tab:shortgpt}). Accuracy collapses with only a
few pruned blocks, and speedup remains limited (often below 1.0$\times$ on
Qwen2.5-0.5B), reflecting the efficiency paradox in Section~\ref{sec:benefits2}. Because
ShortGPT removes attention and FFN together at block granularity, it cannot
exploit the stage- and sub-layer-specific sensitivity that LiteStage targets.
For reference, LiteStage reaches 59.87\% / 1.52$\times$ (TinyLlama, OBQA) and
62.67\% / 1.33$\times$ (Qwen, OBQA) in Table~\ref{tab:multiseed}.

\begin{table}[H]
\centering
\small
\caption{ShortGPT: test accuracy (\%) / end-to-end speedup ($\times$).}
\label{tab:shortgpt}
\resizebox{\linewidth}{!}{%
\begin{tabular}{lccc}
\toprule
\#Pruned blocks & OBQA & CSQA & StrategyQA \\
\midrule
\multicolumn{4}{l}{\emph{TinyLlama-1.1B}} \\
0 (full) & 64.00 / 1.00 & 54.79 / 1.00 & 62.45 / 1.00 \\
2 & 26.80 / 1.06 & 53.48 / 1.10 & 57.64 / 1.09 \\
4 & 13.80 / 1.15 & 45.70 / 1.18 & 2.18 / 1.17 \\
6 & 11.20 / 1.23 & 0.16 / 1.25 & 0.00 / 0.83 \\
8 & 0.00 / 1.04 & 0.00 / 0.75 & 0.00 / 0.46 \\
\midrule
\multicolumn{4}{l}{\emph{Qwen2.5-0.5B}} \\
0 (full) & 70.20 / 1.00 & 64.37 / 1.00 & 65.50 / 1.00 \\
1 & 56.00 / 1.04 & 60.61 / 0.98 & 58.52 / 0.96 \\
2 & 23.40 / 0.98 & 0.98 / 0.95 & 9.61 / 0.86 \\
3 & 0.00 / 1.02 & 0.00 / 0.54 & 0.00 / 0.81 \\
\bottomrule
\end{tabular}}
\end{table}

\subsection{Stage-wise Allocation without Generation Early Exit}
\label{app:noexit}

To isolate the effect of stage-wise allocation, we disable generation early
exit in LiteStage and compare it with AdaSkip on the OBQA \emph{validation} set
using TinyLlama-1.1B (Table~\ref{tab:noexit}). Even without early exit,
LiteStage achieves higher accuracy at comparable speedup (62.2\% vs.\ 61.2\% at
$\sim$1.18$\times$), while AdaSkip collapses to 4.0\% at 1.21$\times$. At lower
speedups, LiteStage improves both accuracy and speedup (64.0\% / 1.10$\times$
vs.\ 62.8\% / 1.03$\times$).

\begin{table}[H]
\centering
\small
\caption{LiteStage without generation early exit vs.\ AdaSkip
(TinyLlama-1.1B, OBQA validation).}
\label{tab:noexit}
\begin{tabular}{lcc}
\toprule
Method & Acc (\%) & Speedup ($\times$) \\
\midrule
Full-layer & 62.40 & 1.000 \\
\midrule
AdaSkip & 62.80 & 1.031 \\
AdaSkip & 62.40 & 1.166 \\
AdaSkip & 61.20 & 1.178 \\
AdaSkip & 4.00 & 1.213 \\
\midrule
LiteStage w/o early exit & 64.00 & 1.100 \\
LiteStage w/o early exit & 62.60 & 1.137 \\
LiteStage w/o early exit & 62.20 & 1.183 \\
\bottomrule
\end{tabular}
\end{table}

\section{Diagnostic Study on Deep Reasoning}
\label{sec:supp_deep_reasoning}

\paragraph{Setups.} To examine the applicability of LiteStage to deep reasoning tasks, we evaluate layer-skipping behavior using Qwen3-1.7B in reasoning mode~\cite{yang2025qwen3}. We use the off-the-shelf model without additional fine-tuning. Evaluation benchmarks include AIME 2025 (AIME 25; mathematics)~\cite{aime2025benchmark}, GPQA-Diamond (GPQA-D; question answering)~\cite{rein2024gpqa}, and LiveCodeBench release v5 (LCB-v5; coding)~\cite{jain2024livecodebench} datasets. Our evaluation largely follows the pipeline provided by QwQ~\cite{qwq32b}\footnote{https://github.com/QwenLM/QwQ}.

Since these benchmarks do not provide validation splits, we first perform "Step 1: Estimate Layer Importance" using the OBQA, CSQA, and StrategyQA datasets, and then compute the average cosine similarities across decoding layers. We treat deep reasoning inference as a single-stage process, which eliminates the need for "Step 2: Search Layer Budget" for stage-wise optimization; accordingly, we apply uniform layer skipping across the entire decoding process. For "Step 3: Generation Early Exit", we observe that the Qwen3-1.7B generally exhibits higher confidence than the models used in short multi-stage reasoning. We therefore increase the confidence threshold to 0.8. Other generation hyperparameters are set to a temperature of 0.6, top-p of 0.95, top-k of 20, and a maximum of 32,768 new tokens.

\paragraph{Layer Skipping.} We apply sub-layer-level skipping (\textit{e.g.}, MHSA or/and FFN), except for the first and last four decoding layers. Layer skipping is applied only during decoding, while the prefill stage is always executed at full depth. We observe that applying layer skipping alone leads to substantial accuracy degradation, even when skipping only one or two layers (\textit{i.e.}, two or four sub-layers). For instance, on AIME 25 and LCB-v5, skipping two layers reduces accuracy from $30.0\% \rightarrow 13.3\%$ and $34.8\% \rightarrow 11.8\%$, respectively. On GPQA-D, skipping a single layer already decreases accuracy from $46.0\% \rightarrow 32.8\%$. These results are summarized in Table~\ref{table:supp_deep_reasoning} (LS; row 1-3).

These findings indicate that the efficiency-accuracy trade-off in deep reasoning fundamentally differs from that in short multi-stage reasoning. We attribute this gap to the intrinsic characteristics of deep reasoning. Unlike short multi-stage reasoning, where intermediate reasoning can be decomposed into relatively independent stages, deep reasoning requires maintaining and refining long-horizon intermediate states over extended decoding steps. In this regime, approximation errors introduced by layer skipping tend to accumulate rather than being corrected in later stages, leading to rapid accuracy collapse.

\begin{figure*}[t]
\centering
\includegraphics[width=\linewidth]{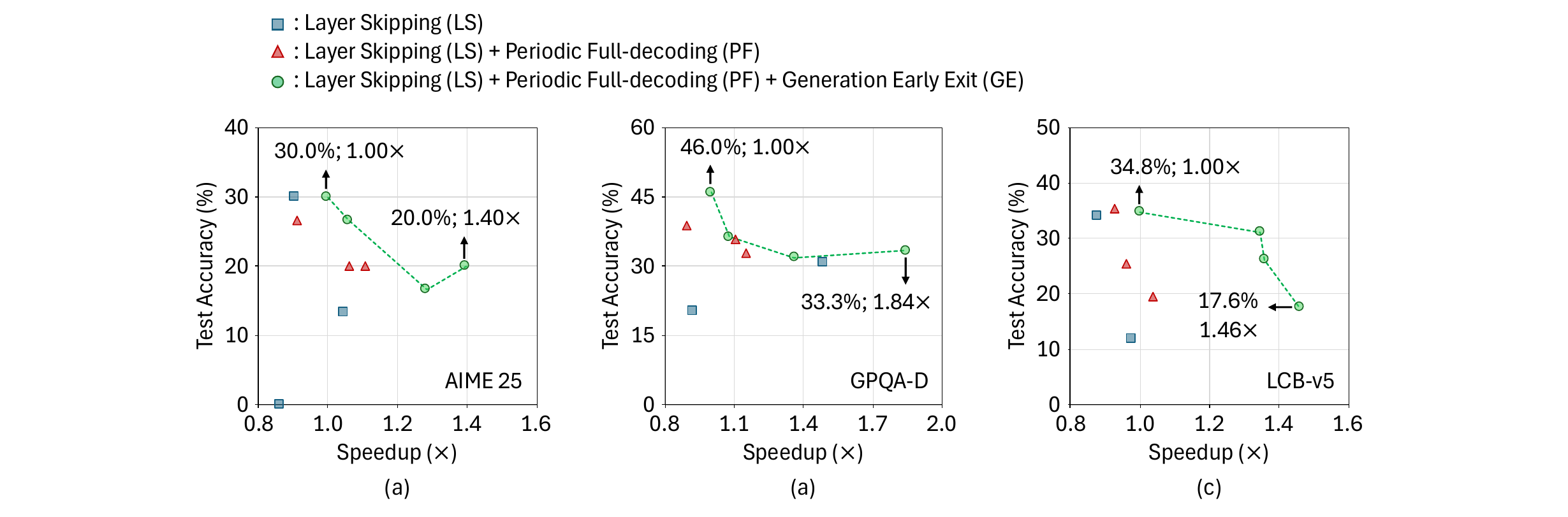}
\caption{\textbf{Deep Reasoning Performance.} Test accuracy and end-to-end speedup are compared between the three configurations (LS, LS+PF, and LS+PF+GE) on the three benchmarks, (a) AIME 25, (b) GPQA-D, and (c) LCB-v5. Results are experimented using Qwen3-1.7B.}

\label{figure:deep_reasoning}
\end{figure*}

\paragraph{Periodic Full-decoding.} Another contributing factor to this discrepancy is the difference in computational flow between multi-stage reasoning and deep reasoning. In multi-stage reasoning, once a stage is completed, the model performs a prefill step that incorporates both the output of the previous stage and the prompt of the current stage. Because our acceleration targets only the decoding phase rather than this prefill step, the model naturally reprocesses the previous stage’s output at full depth, which helps mitigate accumulated approximation errors.

Extending this behavior to deep reasoning models, we introduce periodic full-layer decoding to reduce error accumulation caused by sustained layer skipping. Specifically, we adopt a simple heuristic in which the model decodes the first 1000 tokens using full-layer computation, followed by the next 1000 tokens under layer skipping, and repeats this alternating pattern throughout generation. As shown in Table~\ref{table:supp_deep_reasoning} (LS+PF; row 4-6), this strategy substantially alleviates accuracy degradation. For example, at "\# Skip": 3, accuracy improves from $0.0\% \rightarrow 20.0\%$ on AIME 25, $20.2\% \rightarrow 32.8\%$ on GPQA-D, and $0.0\% \rightarrow 25.5\%$ on LCB-v5.

However, the overall speedup remains marginal, reaching at most 1.15$\times$ (at "\# Skip": 3 on GPQA-D). As observed consistently in our multi-stage reasoning experiments, layer skipping often induces longer generation lengths, preventing per-token speedups from translating into meaningful end-to-end latency reductions.

\paragraph{Generation Early Exit.} Finally, we evaluate the end-to-end speedup achieved by combining generation early exit with layer skipping and periodic full-layer decoding. As shown in Table~\ref{table:supp_deep_reasoning} (LS+PF+GE; row 7-9), this configuration attains speedups of up to 1.40$\times$ on AIME 25, 1.84$\times$ on GPQA-D, and 1.46$\times$ on LCB-v5. These gains are primarily driven by reduced generation lengths, which are also reported in the table.  

Although generation early exit introduces some accuracy degradation at a given number of skipped layers, Figure~\ref{figure:deep_reasoning} demonstrates that incorporating early exit consistently yields superior efficiency–accuracy trade-offs. This result highlights the importance of combining generation early exit with layer skipping to translate per-token computational savings into meaningful end-to-end speedups. 

\paragraph{Discussion.} We find that combining layer skipping with generation early exit improves inference speed. At the same time, \emph{all} evaluated configurations exhibit noticeable accuracy degradation, even under moderate approximation. These observations imply that deep reasoning exhibits a distinct efficiency–accuracy trade-off compared to short multi-stage reasoning. 

This diagnostic study reinforces the design choice of LiteStage. While stage-aware layer skipping is effective for short multi-stage reasoning where stage-level heterogeneity can be exploited to balance latency and accuracy, it is fundamentally mismatched with deep reasoning tasks that demand strong long-horizon consistency. Addressing deep reasoning efficiently likely requires complementary mechanisms, such as the periodic full-decoding, which is however beyond the scope of this work.




\section{Evaluation in Larger Models}
\label{sec:supp_4b_7b}

\begin{figure*}[t]
\centering

\begin{minipage}[t]{0.48\textwidth}
\captionsetup{type=table}

\centering
\renewcommand{\arraystretch}{1.0}
\setlength{\tabcolsep}{1.5pt}
\fontsize{8pt}{8pt}\selectfont
\caption{Layer-skipping results in Qwen2.5-7B on OBQA. "\# Skip (Stg1 / Stg2 / Stg3)" denote the number of skipped layers for Stage 1, Stage 2, and Stage 3. "Acc." and "Dec. Steps" represent the validation accuracy and the total number of decoding steps.}
\label{tab:qwen7b_obqa}
\begin{tabular}{lcccc}
\toprule
\textbf{Model} & \textbf{\# Skip (Stg1 / Stg2 / Stg3)} & \textbf{Acc.} & \textbf{Speedup} & \textbf{Dec. Steps} \\
\midrule
Full-layer & 0 / 0 / 0 & 90.00\% & 1.000$\times$ & 285.2 \\
\midrule
AdaSkip & 2 / 2 / 2 & 88.80\% & 0.974$\times$ & 298.5 \\
AdaSkip & 4 / 4 / 4 & 87.20\% & 0.860$\times$ & 354.0 \\
AdaSkip & 6 / 6 / 6 & 85.80\% & 0.842$\times$ & 346.0 \\
LiteStage & 0 / 2 / 10 & 90.00\% & 1.016$\times$ & 285.3 \\
LiteStage & 0 / 30 / 2 & 89.00\% & 1.748$\times$ & 81.8 \\
LiteStage & 22 / 30 / 10 & 87.00\% & 2.356$\times$ & 39.6 \\
\bottomrule
\end{tabular}

\vspace{0.8em}

\centering
\renewcommand{\arraystretch}{1.0}
\setlength{\tabcolsep}{1.5pt}
\fontsize{8pt}{8pt}\selectfont
\caption{Layer-skipping results in Qwen2.5-7B on CSQA.}
\label{tab:qwen7b_csqa}
\begin{tabular}{lcccc}
\toprule
\textbf{Model} & \textbf{\# Skip (Stg1 / Stg2 / Stg3)} & \textbf{Acc.} & \textbf{Speedup} & \textbf{Dec. Steps} \\
\midrule
Full-layer & 0 / 0 / 0 & 82.98\% & 1.000$\times$ & 279.4 \\
\midrule
AdaSkip & 2 / 2 / 2 & 81.65\% & 1.032$\times$ & 283.7 \\
AdaSkip & 4 / 4 / 4 & 79.77\% & 1.012$\times$ & 301.8 \\
AdaSkip & 6 / 6 / 6 & 79.44\% & 0.972$\times$ & 312.7 \\
LiteStage & 0 / 0 / 2 & 81.65\% & 1.178$\times$ & 236.0 \\
LiteStage & 26 / 0 / 2 & 80.92\% & 1.360$\times$ & 208.4 \\
LiteStage & 28 / 24 / 4 & 80.42\% & 2.347$\times$ & 57.0 \\
\bottomrule
\end{tabular}

\vspace{0.8em}

\centering
\renewcommand{\arraystretch}{1.0}
\setlength{\tabcolsep}{1.5pt}
\fontsize{8pt}{8pt}\selectfont
\caption{Layer-skipping results in Qwen2.5-7B on StrategyQA.}
\label{tab:qwen7b_strategyqa}
\begin{tabular}{lcccc}
\toprule
\textbf{Model} & \textbf{\# Skip (Stg1 / Stg2 / Stg3)} & \textbf{Acc.} & \textbf{Speedup} & \textbf{Dec. Steps} \\
\midrule
Full-layer & 0 / 0 / 0 & 73.80\% & 1.000$\times$ & 248.4 \\
\midrule
AdaSkip & 2 / 2 / 2 & 68.56\% & 0.996$\times$ & 247.6 \\
AdaSkip & 4 / 4 / 4 & 71.18\% & 1.012$\times$ & 264.3 \\
AdaSkip & 6 / 6 / 6 & 74.24\% & 0.972$\times$ & 277.9 \\
AdaSkip & 8 / 8 / 8 & 72.49\% & 0.683$\times$ & 324.5 \\
AdaSkip & 10 / 10 / 10 & 63.76\% & 0.525$\times$ & 499.0 \\
AdaSkip & 12 / 12 / 12 & 67.69\% & 0.466$\times$ & 587.1 \\
LiteStage & 8 / 36 / 12 & 73.36\% & 1.995$\times$ & 94.9 \\
LiteStage & 28 / 36 / 0 & 72.49\% & 4.343$\times$ & 43.0 \\
LiteStage & 32 / 36 / 16 & 72.95\% & 4.444$\times$ & 41.1 \\
\bottomrule
\end{tabular}
\end{minipage}
\hfill
\begin{minipage}[t]{0.48\textwidth}
\captionsetup{type=table}

\centering
\renewcommand{\arraystretch}{1.0}
\setlength{\tabcolsep}{1.5pt}
\fontsize{8pt}{8pt}\selectfont
\caption{Layer-skipping results in Qwen3-4B on OBQA.}
\label{tab:qwen3_4b_obqa}
\begin{tabular}{lcccc}
\toprule
\textbf{Model} & \textbf{\# Skip (Stg1 / Stg2 / Stg3)} & \textbf{Acc.} & \textbf{Speedup} & \textbf{Dec. Steps} \\
\midrule
Full-layer & 0 / 0 / 0 & 89.00\% & 1.000$\times$ & 276.6 \\
\midrule
AdaSkip & 2 / 2 / 2 & 88.00\% & 1.027$\times$ & 284.9 \\
AdaSkip & 4 / 4 / 4 & 87.80\% & 1.004$\times$ & 288.3 \\
AdaSkip & 6 / 6 / 6 & 87.20\% & 0.987$\times$ & 308.8 \\
LiteStage & 2 / 8 / 12 & 88.80\% & 1.165$\times$ & 255.4 \\
LiteStage & 0 / 22 / 2 & 88.60\% & 1.303$\times$ & 242.0 \\
LiteStage & 0 / 54 / 4 & 88.00\% & 2.118$\times$ & 76.3 \\
\bottomrule
\end{tabular}

\vspace{0.8em}

\centering
\renewcommand{\arraystretch}{1.0}
\setlength{\tabcolsep}{1.5pt}
\fontsize{8pt}{8pt}\selectfont
\caption{Layer-skipping results in Qwen3-4B on CSQA.}
\label{tab:qwen3_4b_csqa}
\begin{tabular}{lcccc}
\toprule
\textbf{Model} & \textbf{\# Skip (Stg1 / Stg2 / Stg3)} & \textbf{Acc.} & \textbf{Speedup} & \textbf{Dec. Steps} \\
\midrule
Full-layer & 0 / 0 / 0 & 78.21\% & 1.000$\times$ & 283.8 \\
\midrule
AdaSkip & 2 / 2 / 2 & 76.74\% & 0.985$\times$ & 285.6 \\
AdaSkip & 4 / 4 / 4 & 77.97\% & 0.958$\times$ & 298.6 \\
AdaSkip & 6 / 6 / 6 & 77.48\% & 0.969$\times$ & 287.8 \\
AdaSkip & 8 / 8 / 8 & 76.90\% & 0.996$\times$ & 295.4 \\
AdaSkip & 10 / 10 / 10 & 77.89\% & 1.076$\times$ & 287.8 \\
AdaSkip & 12 / 12 / 12 & 76.00\% & 0.977$\times$ & 316.3 \\
LiteStage & 0 / 52 / 0 & 78.21\% & 2.197$\times$ & 69.0 \\
LiteStage & 0 / 52 / 2 & 77.81\% & 2.232$\times$ & 67.5 \\
LiteStage & 42 / 54 / 6 & 77.47\% & 2.911$\times$ & 38.7 \\
\bottomrule
\end{tabular}

\vspace{0.8em}

\centering
\renewcommand{\arraystretch}{1.0}
\setlength{\tabcolsep}{1.5pt}
\fontsize{8pt}{8pt}\selectfont
\caption{Layer-skipping results in Qwen3-4B on StrategyQA.}
\label{tab:qwen3_4b_strategyqa}
\begin{tabular}{lcccc}
\toprule
\textbf{Model} & \textbf{\# Skip (Stg1 / Stg2 / Stg3)} & \textbf{Acc.} & \textbf{Speedup} & \textbf{Dec. Steps} \\
\midrule
Full-layer & 0 / 0 / 0 & 70.74\% & 1.000$\times$ & 232.4 \\
\midrule
AdaSkip & 2 / 2 / 2 & 69.87\% & 0.880$\times$ & 265.9 \\
AdaSkip & 4 / 4 / 4 & 69.00\% & 0.800$\times$ & 254.0 \\
AdaSkip & 6 / 6 / 6 & 73.36\% & 0.975$\times$ & 249.6 \\
AdaSkip & 8 / 8 / 8 & 66.81\% & 0.879$\times$ & 260.7 \\
AdaSkip & 10 / 10 / 10 & 60.26\% & 0.823$\times$ & 255.3 \\
AdaSkip & 12 / 12 / 12 & 63.76\% & 1.161$\times$ & 231.9 \\
LiteStage & 0 / 36 / 0 & 73.36\% & 2.080$\times$ & 85.0 \\
LiteStage & 0 / 36 / 14 & 71.18\% & 2.107$\times$ & 85.0 \\
LiteStage & 26 / 36 / 0 & 69.40\% & 2.629$\times$ & 67.1 \\
\bottomrule
\end{tabular}
\end{minipage}

\end{figure*}

To evaluate whether LiteStage generalizes beyond small models (${<}$1.1B), we additionally test Qwen3-4B-Instruct-2507 and Qwen2.5-7B-Instruct, following exactly the same multi-stage reasoning workflow used for TinyLlama-1.1B and Qwen2.5-0.5B.

Across OBQA, CSQA, and StrategyQA, LiteStage consistently provides larger end-to-end speedups than both the full-layer baseline and AdaSkip while maintaining comparable accuracy. Tables~\ref{tab:qwen7b_obqa}–\ref{tab:qwen3_4b_strategyqa} report full results including accuracy, end-to-end speedup, and decoding steps.

\paragraph{Results on 7B.} AdaSkip again shows negligible or even negative speedups due to increased generation length (\textit{e.g.}, 0.842$\times$-0.974$\times$ on OBQA, 0.972$\times$-1.032$\times$ on CSQA, and 0.466$\times$-1.012$\times$ on StrategyQA). In contrast, LiteStage achieves substantial acceleration (1.016$\times$-2.356$\times$ on OBQA, 1.178$\times$-2.347$\times$ on CSQA, and 1.995$\times$-4.444$\times$ on StrategyQA) while preserving accuracy close to the full-layer model.

\paragraph{Results on 4B.} A similar trend is observed in Qwen3-4B. AdaSkip provides only marginal speedup (0.987$\times$-1.027$\times$ on OBQA, 0.958$\times$-1.076$\times$ on CSQA, and 0.800$\times$-1.161$\times$ on StrategyQA), whereas LiteStage achieves significantly larger gains (1.165$\times$-2.118$\times$ on OBQA, 2.197$\times$-2.911$\times$ on CSQA, and 2.080$\times$-2.629$\times$ on StrategyQA) with comparable accuracy.

\begin{figure*}[t]
\centering
\includegraphics[width=\linewidth]{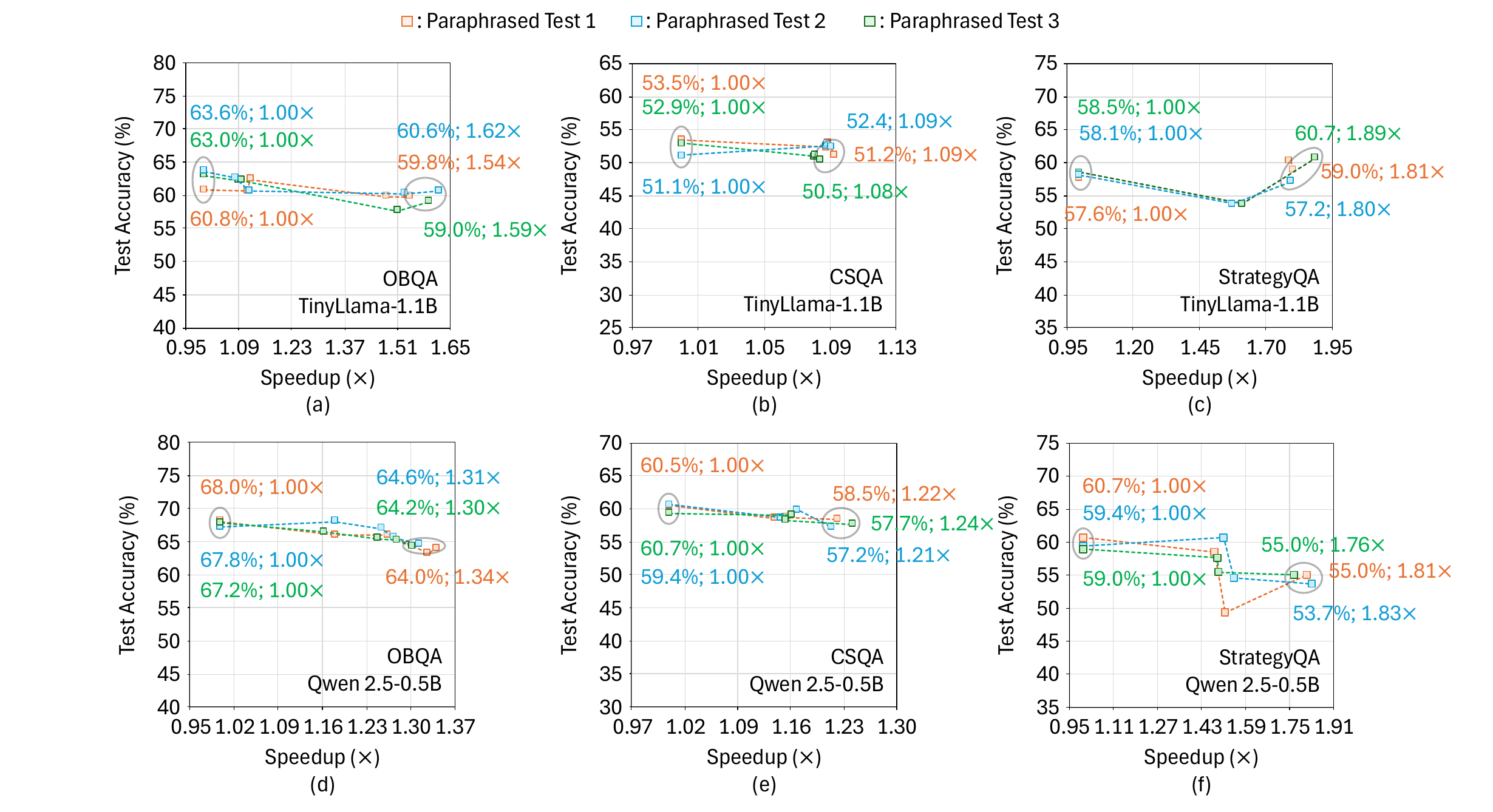}
\caption{\textbf{Evaluation under Mild Distribution Shifts.} For each dataset, we evaluate three paraphrased versions of the original test set using the skip configuration selected on the original validation set without additional tuning. Results are shown for TinyLlama-1.1B and Qwen2.5-0.5B. Across all datasets and paraphrased splits, LiteStage maintains similar accuracy–speedup trade-offs, demonstrating that the validation-selected configuration remains effective under mild distribution shifts.}

\label{figure:distribution_shift}
\end{figure*}

\paragraph{Effect of Model Scale.}
Larger models show stronger robustness to layer skipping and generation early exit. Under aggressive skipping in Stage~2, LiteStage often produces lower output confidence earlier, leading to shorter decoding sequences (${<}$100 steps), yet still maintaining accuracy close to the full-layer model. This suggests that larger models can produce reliable answers with fewer decoding steps once confident tokens are generated.

Overall, these results demonstrate that the effectiveness of LiteStage is not limited to small models. Even for larger 4B and 7B models, the same stage-wise skip strategy continues to provide consistent speedups with minimal accuracy degradation, indicating that LiteStage scales well with model size and can be applied as a general acceleration framework across different model scales.

\section{Evaluation under Distribution Shifts}
\label{sec:supp_dist_shift}
\paragraph{Setups.} To examine whether the validation-selected skip configuration generalizes across mild distribution shifts, we construct additional paraphrased test splits for OBQA, CSQA, and StrategyQA using an external LLM. This experiment tests whether the offline search configuration transfers beyond the original validation/test distribution.

\paragraph{Paraphrased Test Split Generation.} For each dataset, we generate three additional paraphrased versions of the original test set using GPT-4.1-mini. The paraphrasing is applied only to the question text, while the answer choices and ground-truth labels remain unchanged. We instruct the model to produce mild paraphrases that preserve the exact semantics of the original question.

The system prompt used for paraphrasing is shown below:

\begin{quote}
\small
\begin{verbatim}
You rewrite dataset questions.

Task:
Generate 3 mild paraphrases of the question 
text.

Rules:
- Keep meaning exactly the same
- Do NOT change answer meaning
- Do NOT include choices
- Do NOT include explanation
- Do NOT number them
- Output exactly 3 lines
- Each line must be one paraphrase
\end{verbatim}
\end{quote}

\noindent The user prompt template is:

\begin{quote}
\small
\begin{verbatim}
Rewrite the following question into 3 
different mild paraphrases.

Question:
{question}
\end{verbatim}
\end{quote}

\noindent Each original test sample is therefore expanded into three paraphrased variants, resulting in three additional test splits per dataset.

\begin{figure*}[t]
\centering
\includegraphics[width=\linewidth]{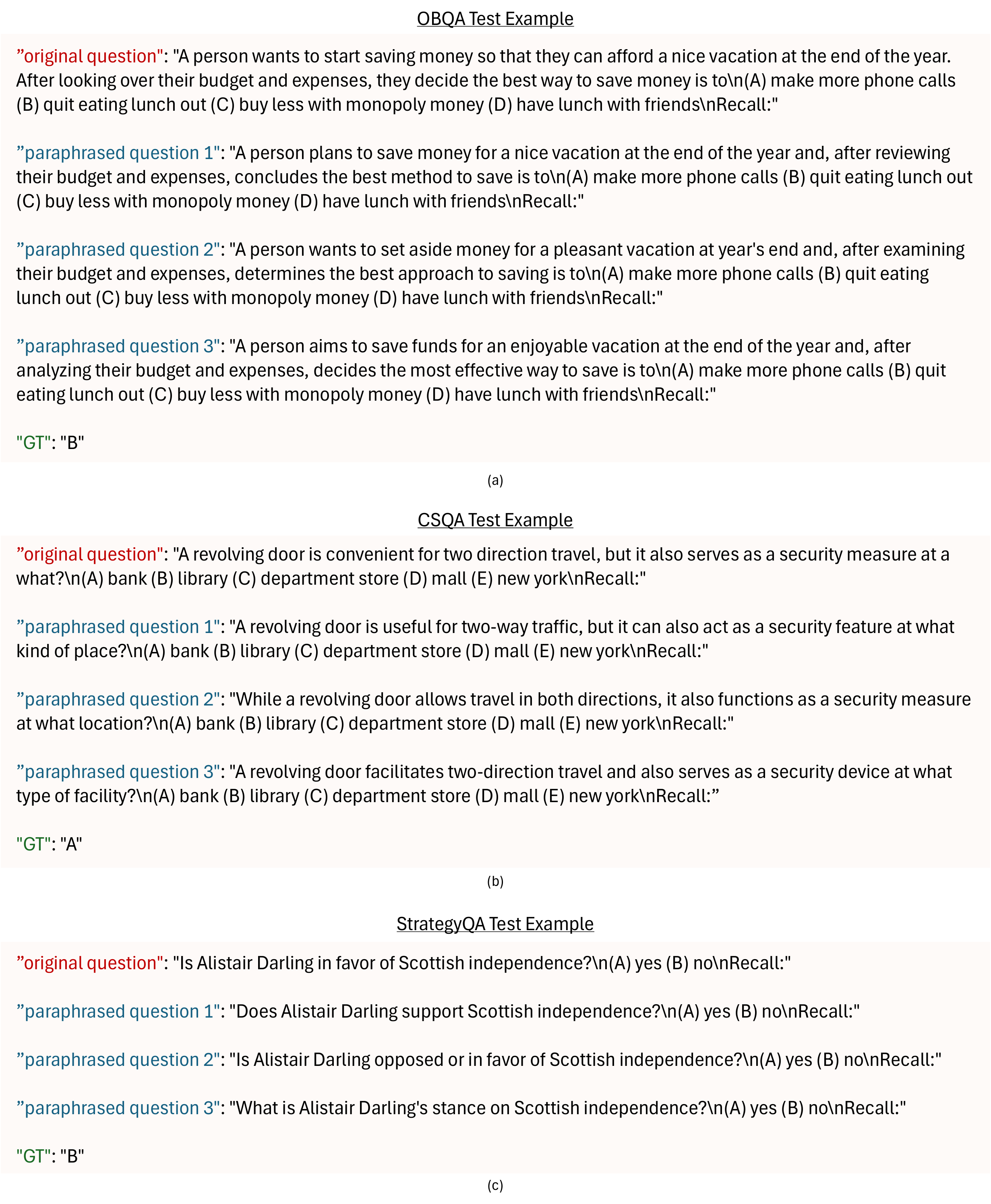}
\caption{\textbf{Example of Paraphrased Test Data.} For each original question in (a) OBQA, (b) CSQA, and (c) StrategyQA, the paraphrases modify only the wording of the question while preserving its semantics, allowing us to test whether the validation-selected skip configuration generalizes beyond the original data distribution.}
\label{figure:supp_paraphrased_example}
\end{figure*}

\noindent\textbf{Construction of Evaluation Splits.} For each dataset (OBQA, CSQA, StrategyQA), we create original test split and paraphrased split 1, 2, and 3. Each paraphrased split contains the same number of samples as the original test set, with identical answer choices and labels, but with paraphrased question text. Examples of the generated paraphrases are shown in Figure~\ref{figure:supp_paraphrased_example}. As illustrated, the paraphrases modify wording while keeping the semantics and answer unchanged.
For instance, the original question:

\begin{quote}
\small
\texttt{A person wants to start saving money so that they can afford a nice vacation at the end of the year. After looking over their budget and expenses, they decide the best way to save money is to ...}
\end{quote}

\noindent is rewritten into semantically equivalent variants such as:

\begin{quote}
\small
\texttt{A person plans to save money for a nice vacation at the end of the year and, after reviewing their budget and expenses, concludes the best method to save is to ...}
\end{quote}

\noindent while preserving the same answer.

\paragraph{Generalization under Mild Distribution Shifts.} Figure~\ref{figure:distribution_shift} shows the results on three paraphrased test splits for OBQA, CSQA, and StrategyQA, where the skip configuration selected on the original validation set is applied without re-tuning. Overall, the accuracy–speedup trade-offs remain similar across paraphrased splits for both TinyLlama-1.1B and Qwen2.5-0.5B, indicating that the offline-selected configuration transfers well under mild distribution shifts.

On OBQA, even in the highest-speedup regime, the results remain close across paraphrased splits.For example, TinyLlama-1.1B achieves 60.6\% at 1.62$\times$, 59.8\% at 1.54$\times$, and 59.0\% at 1.59$\times$, while Qwen2.5-0.5B shows 64.6\% at 1.31$\times$, 64.0\% at 1.34$\times$, and 64.2\% at 1.30$\times$. CSQA shows similar stability, where TinyLlama-1.1B obtains 52.4\% at 1.09$\times$, 51.2\% at 1.09$\times$, and 50.5\% at 1.08$\times$, and Qwen2.5-0.5B achieves 57.2\% at 1.21$\times$, 58.5\% at 1.22$\times$, and 57.7\% at 1.24$\times$.

We observe relatively larger variation on StrategyQA, but the overall trend is still consistent. At the highest speedup point, TinyLlama-1.1B shows 60.1\% at 1.89$\times$, 59.0\% at 1.81$\times$, and 57.2\% at 1.80$\times$, while Qwen2.5-0.5B reports 55.0\% at 1.76$\times$, 55.0\% at 1.81$\times$, and 53.7\% at 1.83$\times$. This larger fluctuation is expected since StrategyQA has only two answer options, making the accuracy more sensitive to small prediction changes compared to multi-choice datasets.

Overall, these results suggest that under the domain-conditioned multi-stage reasoning setting considered in this work, the validation-selected skip configuration generalizes across splits and remains effective under mild distribution shifts, supporting the practicality of the offline search procedure.

%% file: custom.bib
@article{xue2024decompose,
  title={Decompose, analyze and rethink: Solving intricate problems with human-like reasoning cycle},
  author={Xue, Shangzi and Huang, Zhenya and Liu, Jiayu and Lin, Xin and Ning, Yuting and Jin, Binbin and Li, Xin and Liu, Qi},
  journal={Advances in Neural Information Processing Systems},
  volume={37},
  pages={357--385},
  year={2024}
}

@article{dai2024improve,
  title={Improve Student's Reasoning Generalizability through Cascading Decomposed CoTs Distillation},
  author={Dai, Chengwei and Li, Kun and Zhou, Wei and Hu, Songlin},
  journal={arXiv preprint arXiv:2405.19842},
  year={2024}
}

@misc{xu2025llava,
      title={LLaVA-CoT: Let Vision Language Models Reason Step-by-Step}, 
      author={Guowei Xu and Peng Jin and Ziang Wu and Hao Li and Yibing Song and Lichao Sun and Li Yuan},
      year={2025},
      eprint={2411.10440},
      archivePrefix={arXiv},
      primaryClass={cs.CV},
      url={https://arxiv.org/abs/2411.10440}, 
}

@article{piao2024tinythinker,
  title={TinyThinker: Distilling Reasoning through Coarse-to-Fine Knowledge Internalization with Self-Reflection},
  author={Piao, Shengmin and Park, Sanghyun},
  journal={arXiv preprint arXiv:2412.08024},
  year={2024}
}

@article{zhou2024self,
  title={Self-discover: Large language models self-compose reasoning structures},
  author={Zhou, Pei and Pujara, Jay and Ren, Xiang and Chen, Xinyun and Cheng, Heng-Tze and Le, Quoc V and Chi, Ed and Zhou, Denny and Mishra, Swaroop and Zheng, Huaixiu Steven},
  journal={Advances in Neural Information Processing Systems},
  volume={37},
  pages={126032--126058},
  year={2024}
}

@inproceedings{li2024teaching,
  title={Teaching Small Language Models to Reason for Knowledge-Intensive Multi-Hop Question Answering},
  author={Li, Xiang and He, Shizhu and Lei, Fangyu and JunYang, JunYang and Su, Tianhuang and Liu, Kang and Zhao, Jun},
  booktitle={Findings of the Association for Computational Linguistics: ACL 2024},
  pages={7804--7816},
  year={2024}
}

@inproceedings{kim2024llm,
  title={An llm compiler for parallel function calling},
  author={Kim, Sehoon and Moon, Suhong and Tabrizi, Ryan and Lee, Nicholas and Mahoney, Michael W and Keutzer, Kurt and Gholami, Amir},
  booktitle={Forty-first International Conference on Machine Learning},
  year={2024}
}

@inproceedings{he2025adaskip,
  title={Adaskip: Adaptive sublayer skipping for accelerating long-context llm inference},
  author={He, Zhuomin and Yao, Yizhen and Zuo, Pengfei and Gao, Bin and Li, Qinya and Zheng, Zhenzhe and Wu, Fan},
  booktitle={Proceedings of the AAAI Conference on Artificial Intelligence},
  volume={39},
  number={22},
  pages={24050--24058},
  year={2025}
}

@article{liu2024accelerating,
  title={Accelerating inference in large language models with a unified layer skipping strategy},
  author={Liu, Yijin and Meng, Fandong and Zhou, Jie},
  journal={arXiv preprint arXiv:2404.06954},
  year={2024}
}

@article{del2023skipdecode,
  title={Skipdecode: Autoregressive skip decoding with batching and caching for efficient llm inference},
  author={Del Corro, Luciano and Del Giorno, Allie and Agarwal, Sahaj and Yu, Bin and Awadallah, Ahmed and Mukherjee, Subhabrata},
  journal={arXiv preprint arXiv:2307.02628},
  year={2023}
}

@article{men2024shortgpt,
  title={Shortgpt: Layers in large language models are more redundant than you expect},
  author={Men, Xin and Xu, Mingyu and Zhang, Qingyu and Wang, Bingning and Lin, Hongyu and Lu, Yaojie and Han, Xianpei and Chen, Weipeng},
  journal={arXiv preprint arXiv:2403.03853},
  year={2024}
}

@article{raposo2024mixture,
  title={Mixture-of-depths: Dynamically allocating compute in transformer-based language models},
  author={Raposo, David and Ritter, Sam and Richards, Blake and Lillicrap, Timothy and Humphreys, Peter Conway and Santoro, Adam},
  journal={arXiv preprint arXiv:2404.02258},
  year={2024}
}

@article{bae2025mixture,
  title={Mixture-of-recursions: Learning dynamic recursive depths for adaptive token-level computation},
  author={Bae, Sangmin and Kim, Yujin and Bayat, Reza and Kim, Sungnyun and Ha, Jiyoun and Schuster, Tal and Fisch, Adam and Harutyunyan, Hrayr and Ji, Ziwei and Courville, Aaron and others},
  journal={arXiv preprint arXiv:2507.10524},
  year={2025}
}

@article{fan2024not,
  title={Not all layers of llms are necessary during inference},
  author={Fan, Siqi and Jiang, Xin and Li, Xiang and Meng, Xuying and Han, Peng and Shang, Shuo and Sun, Aixin and Wang, Yequan and Wang, Zhongyuan},
  journal={arXiv preprint arXiv:2403.02181},
  year={2024}
}

@article{elhoushi2024layerskip,
  title={LayerSkip: Enabling early exit inference and self-speculative decoding},
  author={Elhoushi, Mostafa and Shrivastava, Akshat and Liskovich, Diana and Hosmer, Basil and Wasti, Bram and Lai, Liangzhen and Mahmoud, Anas and Acun, Bilge and Agarwal, Saurabh and Roman, Ahmed and others},
  journal={arXiv preprint arXiv:2404.16710},
  year={2024}
}

@article{xin2020deebert,
  title={DeeBERT: Dynamic early exiting for accelerating BERT inference},
  author={Xin, Ji and Tang, Raphael and Lee, Jaejun and Yu, Yaoliang and Lin, Jimmy},
  journal={arXiv preprint arXiv:2004.12993},
  year={2020}
}

@article{chen2023ee,
  title={Ee-llm: Large-scale training and inference of early-exit large language models with 3d parallelism},
  author={Chen, Yanxi and Pan, Xuchen and Li, Yaliang and Ding, Bolin and Zhou, Jingren},
  journal={arXiv preprint arXiv:2312.04916},
  year={2023}
}

@article{reasoningknow,
  title={Reasoning Models Know When They're Right: Probing Hidden States for Self-Verification},
  author={Zhang, Anqi and Chen, Yulin and Pan, Jane and Zhao, Chen and Panda, Aurojit and Li, Jinyang and He, He},
  journal={arXiv preprint arXiv:2504.05419},
  year={2025}
}

@article{escot,
  title={Early Stopping Chain-of-thoughts in Large Language Models},
  author={Mao, Minjia and Yin, Bowen and Zhu, Yu and Fang, Xiao},
  journal={arXiv preprint arXiv:2509.14004},
  year={2025}
}

@article{deer,
  title={Dynamic Early Exit in Reasoning Models},
  author={Yang, Chenxu and Si, Qingyi and Duan, Yongjie and Zhu, Zheliang and Zhu, Chenyu and Li, Qiaowei and Lin, Zheng and Cao, Li and Wang, Weiping},
  journal={arXiv preprint arXiv:2504.15895},
  year={2025}
}

@article{entropy,
  title={Entropy After $\langle \texttt{/Think} \rangle$ for reasoning model early exiting},
  author={Wang, Xi and McInerney, James and Wang, Lequn and Kallus, Nathan},
  journal={arXiv preprint arXiv:2509.26522},
  year={2025}
}

@inproceedings{OpenBookQA2018,
 title={Can a Suit of Armor Conduct Electricity? A New Dataset for Open Book Question Answering},
 author={Todor Mihaylov and Peter Clark and Tushar Khot and Ashish Sabharwal},
 booktitle={EMNLP},
 year={2018}
}

@article{talmor2018commonsenseqa,
  title={Commonsenseqa: A question answering challenge targeting commonsense knowledge},
  author={Talmor, Alon and Herzig, Jonathan and Lourie, Nicholas and Berant, Jonathan},
  journal={arXiv preprint arXiv:1811.00937},
  year={2018}
}

@article{geva2021strategyqa,
  title = {{Did Aristotle Use a Laptop? A Question Answering Benchmark with Implicit Reasoning Strategies}},
  author = {Geva, Mor and Khashabi, Daniel and Segal, Elad and Khot, Tushar and Roth, Dan and Berant, Jonathan},
  journal = {Transactions of the Association for Computational Linguistics (TACL)},
  year = {2021},
}

@misc{zhang2024tinyllama,
      title={TinyLlama: An Open-Source Small Language Model}, 
      author={Peiyuan Zhang and Guangtao Zeng and Tianduo Wang and Wei Lu},
      year={2024},
      eprint={2401.02385},
      archivePrefix={arXiv},
      primaryClass={cs.CL}
}

@article{he2024router,
  title={Router-Tuning: A Simple and Effective Approach for Enabling Dynamic-Depth in Transformers},
  author={He, Shwai and Ge, Tao and Sun, Guoheng and Tian, Bowei and Wang, Xiaoyang and Yu, Dong},
  journal={arXiv preprint arXiv:2410.13184},
  year={2024}
}

@article{luo2025adaptive,
  title={Adaptive layer-skipping in pre-trained llms},
  author={Luo, Xuan and Wang, Weizhi and Yan, Xifeng},
  journal={arXiv preprint arXiv:2503.23798},
  year={2025}
}

@inproceedings{luo2025diffskip,
  title={DiffSkip: Differential Layer Skipping in Large Language Models},
  author={Luo, Xuan and Wang, Weizhi and Yan, Xifeng},
  booktitle={Findings of the Association for Computational Linguistics: ACL 2025},
  pages={7221--7231},
  year={2025}
}

@article{qwen2025qwen25technicalreport,
  title        = {Qwen2.5 Technical Report},
  author       = {An Yang and Baosong Yang and Beichen Zhang and Binyuan Hui and Bo Zheng and Bowen Yu and Chengyuan Li and Dayiheng Liu and Fei Huang and Haoran Wei and Huan Lin and Jian Yang and Jianhong Tu and Jianwei Zhang and Jianxin Yang and Jiaxi Yang and Jingren Zhou and Junyang Lin and Kai Dang and Keming Lu and Keqin Bao and Kexin Yang and Le Yu and Mei Li and Mingfeng Xue and Pei Zhang and Qin Zhu and Rui Men and Runji Lin and Tianhao Li and Tianyi Tang and Tingyu Xia and Xingzhang Ren and Xuancheng Ren and Yang Fan and Yang Su and Yichang Zhang and Yu Wan and Yuqiong Liu and Zeyu Cui and Zhenru Zhang and Zihan Qiu},
  journal      = {arXiv preprint arXiv:2412.15115v2},
  year         = {2024},
  url          = {https://arxiv.org/abs/2412.15115v2},
  doi          = {10.48550/arXiv.2412.15115}
}

@inproceedings{yang2025markov,
  title={Markov chain of thought for efficient mathematical reasoning},
  author={Yang, Wen and Liao, Minpeng and Fan, Kai},
  booktitle={Proceedings of the 2025 Conference of the Nations of the Americas Chapter of the Association for Computational Linguistics: Human Language Technologies (Volume 1: Long Papers)},
  pages={7132--7157},
  year={2025}
}

@inproceedings{wang2025stepwise,
  title={Stepwise informativeness search for improving llm reasoning},
  author={Wang, Siyuan and Zhao, Enda and Ren, Xiang},
  booktitle={Proceedings of the 2025 Conference on Empirical Methods in Natural Language Processing},
  pages={25291--25309},
  year={2025}
}

@article{chen2024not,
  title={Do not think that much for 2+ 3=? on the overthinking of o1-like llms},
  author={Chen, Xingyu and Xu, Jiahao and Liang, Tian and He, Zhiwei and Pang, Jianhui and Yu, Dian and Song, Linfeng and Liu, Qiuzhi and Zhou, Mengfei and Zhang, Zhuosheng and others},
  journal={arXiv preprint arXiv:2412.21187},
  year={2024}
}

@inproceedings{hongru2025self,
  title={Self-reasoning language models: Unfold hidden reasoning chains with few reasoning catalyst},
  author={Hongru, WANG and Cai, Deng and Zhong, Wanjun and Huang, Shijue and Pan, Jeff Z and Liu, Zeming and Wong, Kam-Fai},
  booktitle={Workshop on Reasoning and Planning for Large Language Models},
  year={2025}
}

@article{jin2024impact,
  title={The impact of reasoning step length on large language models},
  author={Jin, Mingyu and Yu, Qinkai and Shu, Dong and Zhao, Haiyan and Hua, Wenyue and Meng, Yanda and Zhang, Yongfeng and Du, Mengnan},
  journal={arXiv preprint arXiv:2401.04925},
  year={2024}
}

@inproceedings{rein2024gpqa,
  title={Gpqa: A graduate-level google-proof q\&a benchmark},
  author={Rein, David and Hou, Betty Li and Stickland, Asa Cooper and Petty, Jackson and Pang, Richard Yuanzhe and Dirani, Julien and Michael, Julian and Bowman, Samuel R},
  booktitle={First Conference on Language Modeling},
  year={2024}
}

@article{jain2024livecodebench,
  title={Livecodebench: Holistic and contamination free evaluation of large language models for code},
  author={Jain, Naman and Han, King and Gu, Alex and Li, Wen-Ding and Yan, Fanjia and Zhang, Tianjun and Wang, Sida and Solar-Lezama, Armando and Sen, Koushik and Stoica, Ion},
  journal={arXiv preprint arXiv:2403.07974},
  year={2024}
}

@misc{aime2025benchmark,
  title        = {AIME 2025: Competition Problems for Mathematical Reasoning Evaluation},
  author       = {{Mathematical Association of America}},
  year         = {2025},
  note         = {Used as a benchmark for evaluating mathematical reasoning in large language models}
}

@article{yang2025qwen3,
  title={Qwen3 technical report},
  author={Yang, An and Li, Anfeng and Yang, Baosong and Zhang, Beichen and Hui, Binyuan and Zheng, Bo and Yu, Bowen and Gao, Chang and Huang, Chengen and Lv, Chenxu and others},
  journal={arXiv preprint arXiv:2505.09388},
  year={2025}
}

@misc{qwq32b,
    title = {QwQ-32B: Embracing the Power of Reinforcement Learning},
    url = {https://qwenlm.github.io/blog/qwq-32b/},
    author = {Qwen Team},
    month = {March},
    year = {2025}
}
